\documentclass[10pt,twocolumn,letterpaper]{article}

\usepackage[pagenumbers]{cvpr} % To force page numbers, e.g. for an arXiv version

\usepackage{graphbox}
\usepackage[dvipsnames]{xcolor}
\usepackage{paralist}
\usepackage{enumitem}
\usepackage{longtable}
\usepackage{multirow}
\usepackage[accsupp]{axessibility}
\newcommand{\Paragraph}[1]
{\vspace{1mm} \noindent\textbf{#1}}

\newcommand\NAME{\texttt{DMP}\xspace}
\newcommand\CODE{\small\url{https://github.com/shinying/dmp}}

\definecolor{cvprblue}{rgb}{0.21,0.49,0.74}
\usepackage[pagebackref,breaklinks,colorlinks,linkcolor=cvprblue,citecolor=cvprblue,urlcolor=cvprblue]{hyperref}
\usepackage{comment}

\def\paperID{***} % *** Enter the Paper ID here
\def\confName{CVPR}
\def\confYear{2024}

\title{Exploiting Diffusion Prior for Generalizable Dense Prediction}

\author{
Hsin-Ying Lee\textsuperscript{1}
\qquad
Hung-Yu Tseng\textsuperscript{2}
\qquad
Hsin-Ying Lee\textsuperscript{3}
\qquad
Ming-Hsuan Yang\textsuperscript{1,4}
\\ \\
\textsuperscript{1}University of California, Merced
\qquad
\textsuperscript{2}Meta
\qquad
\textsuperscript{3}Snap Research
\qquad
\textsuperscript{4}Yonsei University
}

\begin{document}

\twocolumn[{
\renewcommand\twocolumn[1][]{#1}%
\maketitle
\vspace{-3mm}
\centering
\includegraphics[width=\linewidth]{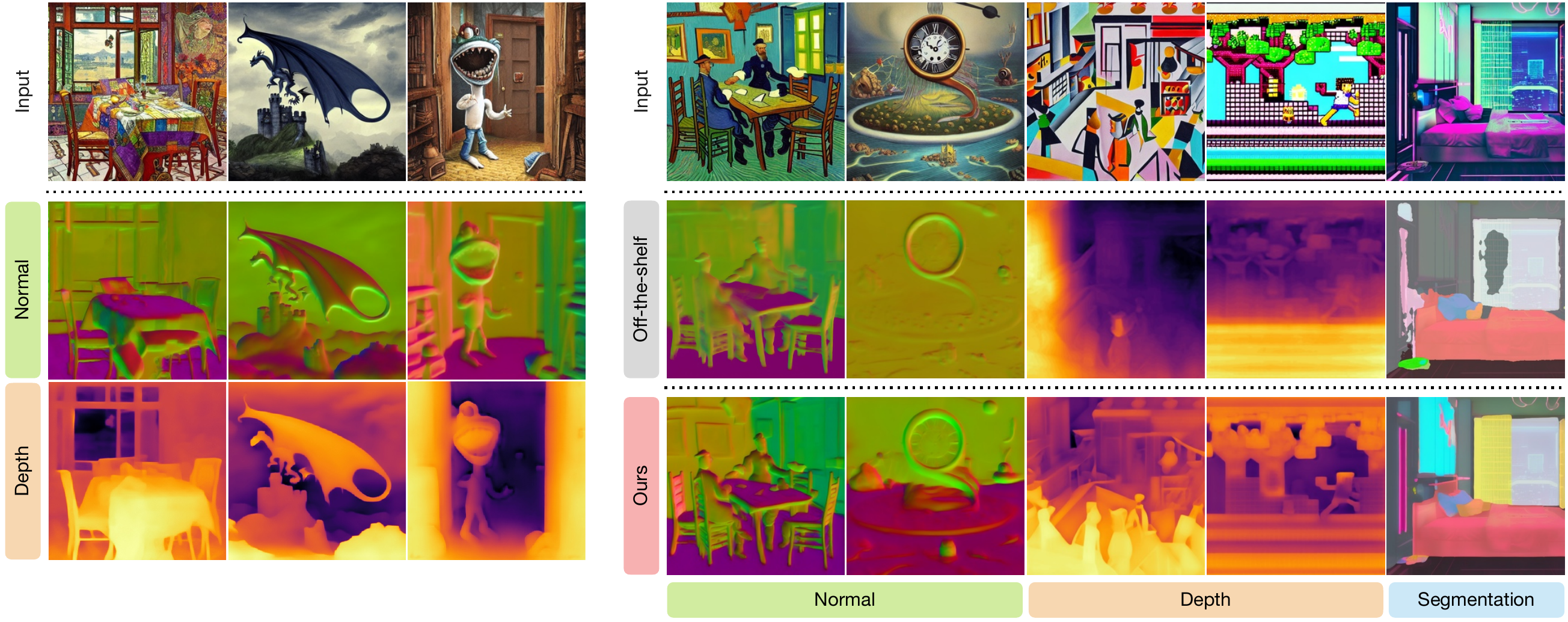}
\vspace{-5mm}
\captionof{figure}{\textbf{Generalized dense prediction}. 
(\textit{left}) We leverage the pre-trained text-to-image diffusion model ~\cite{Rombach_2022_CVPR}  as a \emph{prior} for various dense prediction tasks.
(\textit{right}) With only a small amount of labeled training data in a \emph{limited} domain (i.e., 10K bedroom images with labels) for each task, our method performs favorably against SOTA predictors~\cite{Kar_2022_CVPR, zoedepth, EVA02} on \emph{arbitrary} images.
}
\label{fig:teaser}
\vspace{12pt}
}]

% \maketitle
\begin{abstract}
\vspace{-2mm}
Contents generated by recent advanced Text-to-Image (T2I) diffusion models are sometimes too imaginative for existing off-the-shelf dense predictors to estimate due to the immitigable domain gap.
We introduce \NAME, a pipeline utilizing pre-trained T2I models as a prior for dense prediction tasks. 
To address the misalignment between deterministic prediction tasks and stochastic T2I models, we reformulate the diffusion process through a sequence of interpolations, establishing a deterministic mapping between input RGB images and output prediction distributions.
To preserve generalizability, we use low-rank adaptation to fine-tune pre-trained models. 
Extensive experiments across five tasks, including 3D property estimation, semantic segmentation, and intrinsic image decomposition, showcase the efficacy of the proposed method.
Despite limited-domain training data, the approach yields faithful estimations for arbitrary images, surpassing existing state-of-the-art algorithms. The code is available at \CODE.
\vspace{-3mm}
\end{abstract}   
\section{Introduction}
\label{sec:intro}

Text-to-image~(T2I) diffusion models~\cite{Rombach_2022_CVPR, saharia2022photorealistic, gu2023matryoshka, dai2023emu} have achieved unprecedented progress on text-guided image generation, producing highly imaginative and realistic images from diverse and free-from textual descriptions.
These advancements open up a new era of AI-aided content creation with applications spanning various domains~\cite{hertz2023prompttoprompt,parmar2023zero,shi2023dragdiffusion,Takagi2022brain,avrahami2022blended,ruiz2023dreambooth,rahman2023make}.
However, the creativity of images generated by T2I models poses challenges for off-the-shelf dense (e.g., depth, normal, segmentation) prediction methods~\cite{zoedepth, Kar_2022_CVPR, EVA02} due to the domain gap. 
For example, the ZoeDepth~\cite{zoedepth} approach fails to accurately predict the depth of the cubism painting shown in the sixth column in Figure~\ref{fig:teaser}.
Such property predictions are vital for understanding high-level semantics of generated contents and can facilitate various downstream applications such as 3D imaging~\cite{shih20203d} and relighting~\cite{yeh2022photoscene}.

Existing dense prediction models are typically trained on ``real-world'' images regardless of the training dataset scale.
While these models aim for generalization, bridging the domain gap between real-world and T2I-generated images remains challenging, as we demonstrate in the right-hand side of Figure~\ref{fig:teaser}. 
A potential solution is to take advantage of the inherent generalizability of pre-trained T2I models, for example, by formulating dense prediction as image-to-image translation. However, while several recent efforts have been made to solve various image-to-image translations with pre-trained T2I models~\cite{parmar2023zero,hertz2023prompttoprompt,Brooks_2023_CVPR}, we show in Section~\ref{sec:experiments} that these methods are not directly applicable.

Leveraging pre-trained T2I models as a prior for dense prediction is challenging for two reasons.
First, most dense prediction tasks are inherently \emph{deterministic}, posing difficulties when adapting a pre-trained T2I model designed for stochastic text-to-image generation. 
Second, it is crucial to strike a balance between learning target tasks and retaining the inherent generalizability of pre-trained T2I models.
In other words, learned dense predictors should generalize to arbitrary images from the training data in a limited domain.

In this paper, we propose \NAME (\textbf{D}iffusion \textbf{M}odels as \textbf{P}riors) to leverage the pre-trained T2I model~\cite{Rombach_2022_CVPR} as a prior for generalized dense prediction.
To resolve the determinism-stochasticity misalignment, we introduce a deterministic mapping between the input RGB images and output prediction distributions.
Specifically, we reformulate the diffusion process as a chain of interpolations between input RGB images and their corresponding output signals, where the importance of input images gradually increases over the diffusion process.
The reverse diffusion (i.e., known as denoising or generation in original T2I) process becomes a series of transformations that progressively synthesize desired output signals from input images.
Without randomization, such as Gaussian noise imposed, the mapping is entirely deterministic.
In addition, to retain the generalizability of the pre-trained T2I model while learning target tasks, we use low-rank adaptation~\cite{hu2021lora} to fine-tune the pre-trained model with the aforementioned deterministic diffusion process for each dense prediction task.
Figure~\ref{fig:teaser} demonstrates the generalization ability of the proposed method on the deterministic normal, depth, and segmentation prediction problems.

We conduct extensive quantitative and qualitative experiments on five dense prediction tasks to evaluate the proposed \NAME approach: 3D property estimation (depth, normal), semantic segmentation, and intrinsic image decomposition (albedo, shading).
We show that with only a small amount of limited-domain training data (i.e., 10K bedroom training images with labels), the proposed method can provide faithful estimations of the in-domain and unseen images, especially those that the existing SOTA algorithms struggle to handle effectively.
We summarize the contributions as follows:
\begin{compactitem}
    \item We propose \NAME, an approach leveraging the pre-trained T2I model as a prior for dense prediction tasks.
    \item We design an image-to-prediction diffusion process that adapts the stochastic T2I model for deterministic dense prediction problems.
    \item We use five dense prediction tasks to validate that the proposed method obtains faithful estimation on arbitrary images despite training with a small amount of data in a limited domain.
\end{compactitem}
\section{Related Work}
\label{sec:related}

\Paragraph{Diffusion models.} Diffusion models estimate a target data distribution by modeling the transition from a noisy version of the distribution~\cite{sohl_2015_ICML, ddpm}.
Recently, diffusion models have shown unprecedented quality in the text-to-image~\cite{Rombach_2022_CVPR,saharia2022photorealistic} setting by training on large-scale datasets~\cite{schuhmann2022laionb}.
The advancements unleash various text-guided image manipulation applications~\cite{hertz2023prompttoprompt,parmar2023zero,shi2023dragdiffusion,avrahami2022blended,rahman2023make,cheng2023adaptively,Brooks_2023_CVPR,visii}.
The stochasticity in the generation process, a preferred property in most applications, derives from the initial noise and additional noise added during the denoising process. 
However, the dense prediction problems are usually deterministic.
Some recent methods adopt deterministic sampling algorithms~\cite{song2021denoising, edm, dpm-solver} and reformulate the diffusion and generative process by $\alpha$-blending and de-blending \cite{alphablend}.
Though the mapping between initial noise and outputs in target distributions is deterministic, the correlation between each noise and output sample is stochastic. If adapted to deterministic tasks, the model may generate high-quality outputs unfaithful to input images.

\Paragraph{Image-to-image translation.} The task aims to learn a mapping between two visual domains. 
Early efforts~\cite{zhu2017unpaired,isola2017image,lee2018diverse,zhu2017toward,park2019SPADE,chen2022vector} mostly make use of generative adversarial networks and cycle consistency loss to learn the translation from scratch. 
With StyleGAN~\cite{karras2019style,karras2020analyzing} showing high-quality synthesis on certain categories, some methods~\cite{tov2021designing,richardson2021encoding} seek to achieve image-to-image translation using pre-trained StyleGAN by training an additional encoder. 
However, the translation is limited to certain categories. 
Recently, following the success of large-scale text-to-image models, attempts have been made to perform image-to-image translation with pre-trained diffusion models~\cite{hertz2023prompttoprompt,parmar2023zero,Brooks_2023_CVPR,Tumanyan_2023_CVPR}.
These methods, however, are not directly applicable to the dense prediction task. 

\Paragraph{Fine-tuning text-to-image diffusion models.} 
In addition to image-to-image translation, pre-trained diffusion models are adopted to take additional modalities~\cite{Zhang_2023_ICCV,li2023gligen}, be customized on certain objects~\cite{gal2023an, Ruiz_2023_CVPR} or styles~\cite{Zhang_2023_CVPR}, and synthesize videos~\cite{mahapatra2023synthesizing,guo2023animatediff,geyer2023tokenflow}. 
These methods train additional zero-initialized layers~\cite{Zhang_2023_ICCV}, manipulate attention modules~\cite{geyer2023tokenflow,li2023gligen}, learn a token embedding~\cite{gal2023an, Zhang_2023_CVPR}, or learn parameter offsets with low-rank matrices~\cite{hu2021lora}.
In this work, we adopt the low-rank adaptation~\cite{hu2021lora} to fine-tune only parameter offsets of attention layers.

\Paragraph{Generative prior for dense prediction.} Prior work has leveraged pre-trained generative models as a prior for other tasks, such as representation learning~\cite{donahue2019large}, as the latent features of pre-trained generative models are found to be rich in semantics~\cite{Tritrong_2021_CVPR, Xu2021generative, Xu2021linear}.
\citet{bhattad2023stylegan} manipulates style latents of StyleGAN~\cite{Karras_2019_CVPR} and reveals its learned ability to estimate image properties, but the generalizability is limited. Some works reuse~\cite{baranchuk2022labelefficient, Xu_2023_CVPR, Zhao_2023_ICCV, gong2023prompting} or merge~\cite{tian2023diffuse} latent features of pre-trained denoising U-Nets to perform segmentation and depth estimation. Others~\cite{wang2023incontext, Wang_2023_CVPR} transform generation models into multi-task generalists by standardizing the outputs of tasks as images.
In this work, instead of developing a specific approach for a particular task, we focus on analyzing the potential of pre-trained models as a prior for general dense prediction through a universal transferring framework.
\section{Method}

Our goal is to leverage the pre-trained T2I diffusion model as a prior to learn a dense prediction task from a set of labeled training data $\mathcal{D}=\{(x^i, y^i)\}^n_{i=1}$, where $x^i$ denotes the input image, $y^i$ indicates the corresponding output (e.g., depth map), and $x^i, y^i \in \mathbb{R}^{H\times W\times 3}$.
We first describe the text-to-diffusion models used as the prior in Section~\ref{sec:method_t2i}.
Then we introduce the proposed \NAME approach in Section~\ref{sec:method_ours}.

\subsection{Text-to-Image Diffusion Models}
\label{sec:method_t2i}
We use the pre-trained T2I latent diffusion model~\cite{Rombach_2022_CVPR} as the prior.
It consists of an autoencoder and a U-Net.
The autoencoder converts between an image $y \in \mathbb{R}^{H\times W\times 3}$ and its latent feature $\Tilde{y} \in \mathbb{R}^{h\times\ w \times c}$, where $(h, w) = (H/8, W/8)$ and $c$ represents the channel size of latent features.
Since we do not modify the autoencoder in the proposed approach, we use $y$ to represent the latent feature $\Tilde{y}$ to simplify the annotation in this paper.

The U-Net model takes as input a text description and learns to reverse the following diffusion process that gradually turns the image $y$ into noise map $y_T$:
\begin{equation}
y_t=\sqrt{\bar{\alpha}_t}y + \sqrt{1 - \bar{\alpha}_t}\epsilon_t\hspace{5mm}\epsilon\sim\mathcal{N}(\mathbf{0}, \mathbf{I}),
\label{eq:diffusion}
\end{equation}
where $t=[1,\cdots,T]$, and $\bar{\alpha}_t$ is the noise schedule~\cite{ddpm}.

\subsection{Leveraging Diffusion Prior}
\label{sec:method_ours}

There are two challenges to leverage (\ie, fine-tune) the pre-trained T2I approach for estimating the pixel-level output (\eg, depth) $y$ from an input image $x$: 1) determinism-stochasticity misalignment and 2) generalizability.
We introduce the solutions to tackle these two issues as follows.

\begin{figure}
    \centering
    \includegraphics[width=\linewidth]{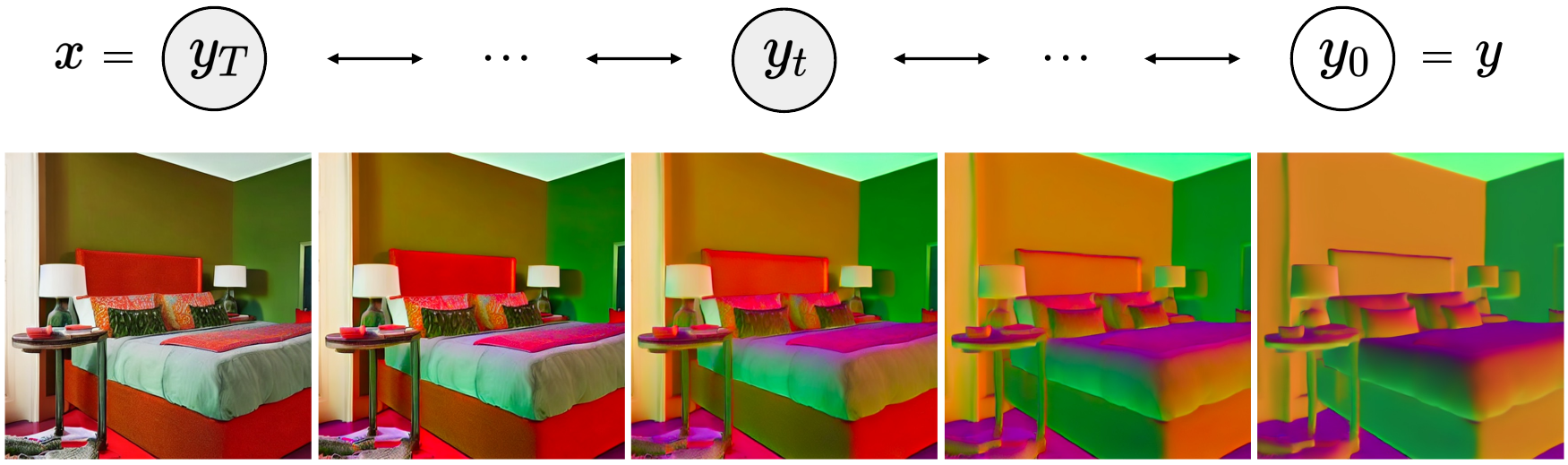}
    \caption{\textbf{Deterministic diffusion process.} We formulate the diffusion process as a chain of interpolations between an input image $x$ and output $y$. 
    The U-Net model is fine-tuned to gradually transform the input $x$ to the desired dense prediction $y$.
    }
    \label{fig:method}
\end{figure}
\begin{figure*}
    \centering
    \includegraphics[width=\linewidth]{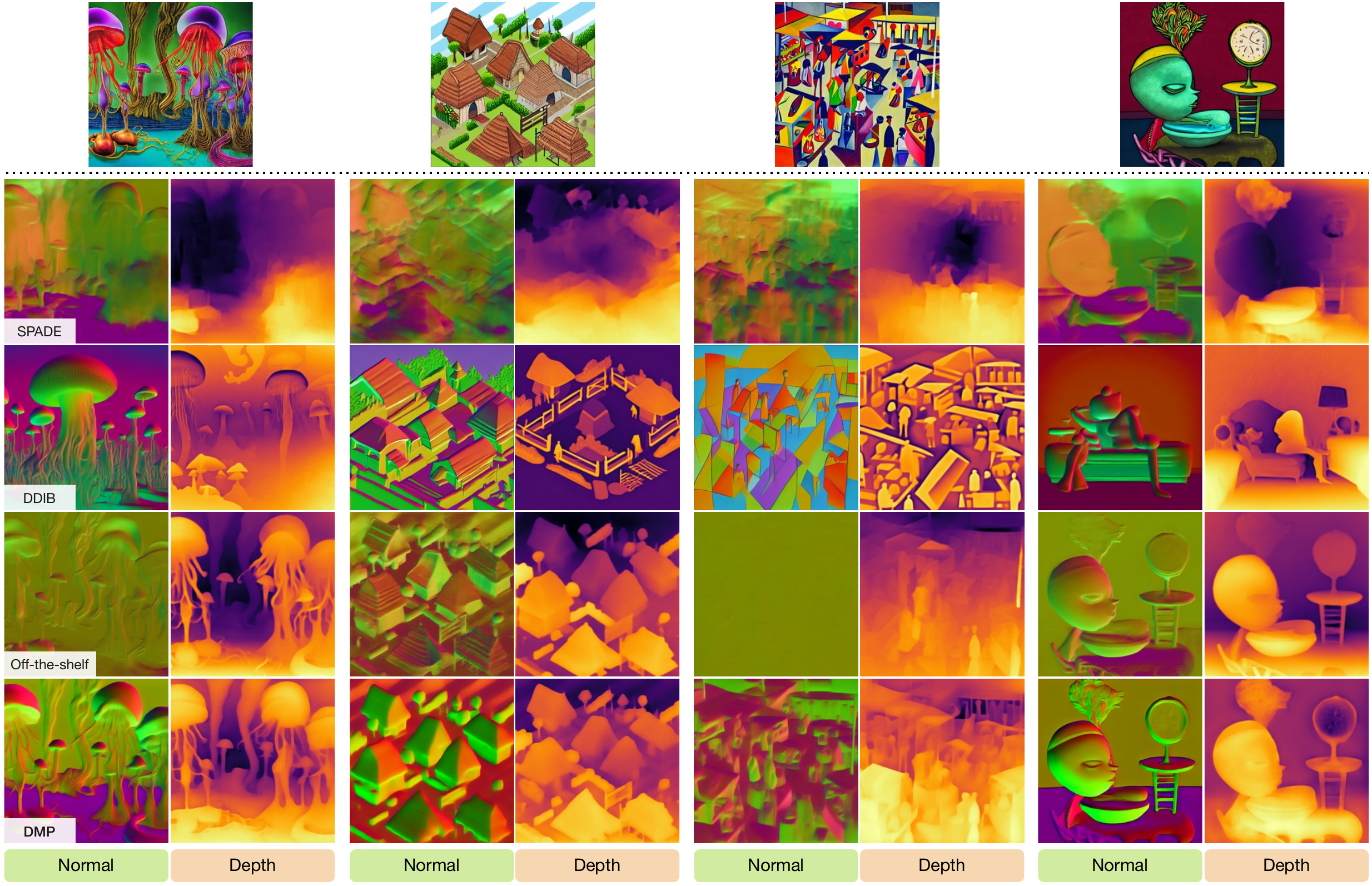}
    \vspace{-2mm}
    \caption{\textbf{3D property estimation of \emph{arbitrary} input images.} The first row shows the input images, while the remaining rows present the normals and depth estimated by different approaches. The proposed \NAME method gives faithful estimation, even on the images where the off-the-shelf~\cite{Kar_2022_CVPR,zoedepth} schemes fail to handle.}
    \vspace{-2mm}
    \label{fig:superood}
\end{figure*}

\Paragraph{Deterministic diffusion.}
The diffusion process described in Eq. \eqref{eq:diffusion} is designed specifically for stochastic image generation.
However, the mapping between input images and outputs in dense prediction problems is typically deterministic.
We observe that directly applying the diffusion process in Eq. \eqref{eq:diffusion} to the dense prediction tasks introduces unnecessary variation in outputs that leads to apparent artifacts.
Therefore, we use the blending strategy~\cite{alphablend,liu2023i2sb} and re-design the diffusion process as follows.
Instead of converting from noise maps to images in the conventional T2I method, the diffusion process in our \NAME approach directly maps between the input image $x$ and output $y$.
As illustrated in Figure~\ref{fig:method}, the proposed diffusion process is formulated as
\begin{equation}
y_t=\sqrt{\bar{\alpha}_t}y + \sqrt{1 - \bar{\alpha}_t}x\hspace{5mm}t=[1,\cdots,T].
\label{eq:dp_diffusion}
\end{equation}
As we can see from Figure~\ref{fig:method} and Eq. \eqref{eq:dp_diffusion}, the proposed scheme gradually increases the weight of the input image $x$ over the diffusion process. 
This can be considered as progressively morphing the output $y$ into the input image $x$ via interpolation.
As a result, we can fine-tune the U-Net model to reverse the diffusion process that interactively ``de-morphs'' the input image $x$ and gives the final prediction $y$.

\Paragraph{Parameterization.}
We explore various parameterizations for the U-Net model to make different predictions.
As discussed in Section~\ref{sec:exp_ablation}, we empirically find that the v-prediction~\cite{salimans2022progressive} works well for the dense prediction tasks.
Specifically, the U-Net model $v_\theta$ is fine-tuned using the following mean square loss:
\begin{equation}
L_\mathrm{DMP}=\mathbb{E}_{(x,y),t}\big[\|(\sqrt{\bar{\alpha}_t}x-\sqrt{1-\bar{\alpha}_t}y) - v_\theta(y_t, t)\|^2_2\big],
\label{eq:loss}
\end{equation}
where $v_\theta(y_t, t)$ is the U-Net prediction.
The reverse diffusion process can then be formulated as
\begin{equation}
\begin{aligned}
y_{t-1}&=\sqrt{\bar{\alpha}_{t-1}}(\sqrt{\bar{\alpha}_t}y_t - \sqrt{1 - \bar{\alpha}_t}v_\theta(y_t, t))\\
&+ \sqrt{1 - \bar{\alpha}_{t-1}}x\hspace{7mm}t=[T,\cdots,1],
\end{aligned}
\end{equation}
where $y_T=x$ and $y_0$ is the desired output.

\Paragraph{U-Net fine-tuning.}
To learn the target tasks while retaining the inherent generalization ability of the pre-trained T2I model, we use the low-rank approximation~\cite{hu2021lora} scheme to fine-tune all the attention layers in the U-Net model to minimize the objective described in Eq. \eqref{eq:loss}.
\section{Experimental Results}
\label{sec:experiments}

\subsection{Experiment Setup}
\label{sec:exp_setup}

We evaluate the proposed \NAME approach using five dense prediction tasks, including 3D property estimation (i.e., depth and normals), semantic segmentation, and intrinsic image decomposition (i.e., albedo and shading).

\begin{table*}[t]
    \centering \footnotesize
    \vspace{-2mm}
    \caption{\textbf{Quantitative comparisons on 3D property estimation.} We compute the metrics using the estimated results and the pseudo ground truth generated by the off-the-shelf predictors.}
    \vspace{-2mm}
    \begin{tabular}{lcccccccccc}
        \toprule
         & \multicolumn{4}{c}{Normal} & \multicolumn{6}{c}{Depth} \\
         \cmidrule(lr){2-5} \cmidrule(lr){6-11}
         & \multicolumn{2}{c}{In-domain} & \multicolumn{2}{c}{Out-of-domain} & \multicolumn{3}{c}{In-domain} & \multicolumn{3}{c}{Out-of-domain} \\
         \cmidrule(lr){2-3} \cmidrule(lr){4-5} \cmidrule(lr){6-8} \cmidrule(lr){9-11}
         & L1$\downarrow$ & Ang$\downarrow$ & L1$\downarrow$ & Ang$\downarrow$ & REL$\downarrow$ & $\delta\uparrow$ & RMSE$\downarrow$ & REL$\downarrow$ & $\delta\uparrow$ & RMSE$\downarrow$ \\
        \midrule
         SPADE \cite{Park_2019_CVPR}  & \underline{0.0708} & \underline{0.1635} & \underline{0.1268} & \underline{0.2833} & \underline{0.2132} & 0.4961 & \underline{0.1379} & \underline{0.3587} & \underline{0.3190} & \underline{0.2554} \\
         DRIT++ \cite{DRIT_plus} & 0.0784 & 0.1723 & 0.1350 & 0.3006 & 0.3792 & 0.2458 & 0.2134 & 0.4374 & 0.2585 & 0.3212 \\
         \midrule
         SDEdit \cite{meng2022sdedit}   & 0.2599 & 0.5087 & 0.2675 & 0.5293 & 0.4656 & 0.3533 & 0.3240 & 0.6640 & 0.2495 & 0.3382 \\
         DDIB \cite{su2023dual}         & 0.1849 & 0.4210 & 0.2271 & 0.4847 & 0.3087 & \underline{0.5130} & 0.2367 & 0.6275 & 0.2788 & 0.3120 \\
         IP2P (hard) \cite{Brooks_2023_CVPR}    & 0.3017 & 0.5468 & 0.3168 & 0.5757 & 0.4834 & 0.3235 & 0.3358 & 0.6450 & 0.2252 & 0.3461 \\
         IP2P (learned) \cite{Brooks_2023_CVPR} & 0.3550 & 0.7181 & 0.3397 & 0.6836 & 0.3965 & 0.3302 & 0.3494 & 0.5182 & 0.2664 & 0.3261 \\
         VISII \cite{visii}                     & 0.2081 & 0.4386 & 0.2448 & 0.4895 & 0.3498 & 0.4405 & 0.2912 & 0.5364 & 0.2855 & 0.3181 \\
         \midrule
         \NAME    & \textbf{0.0514} & \textbf{0.1156} & \textbf{0.0872} & \textbf{0.1886} & \textbf{0.1072} & \textbf{0.8861} & \textbf{0.1020} & \textbf{0.2117} & \textbf{0.6395} & \textbf{0.1360} \\
        \bottomrule
    \end{tabular}
    \label{tab:normal-depth}
\end{table*}
\begin{figure*}
    \centering
    \includegraphics[width=0.99\linewidth]{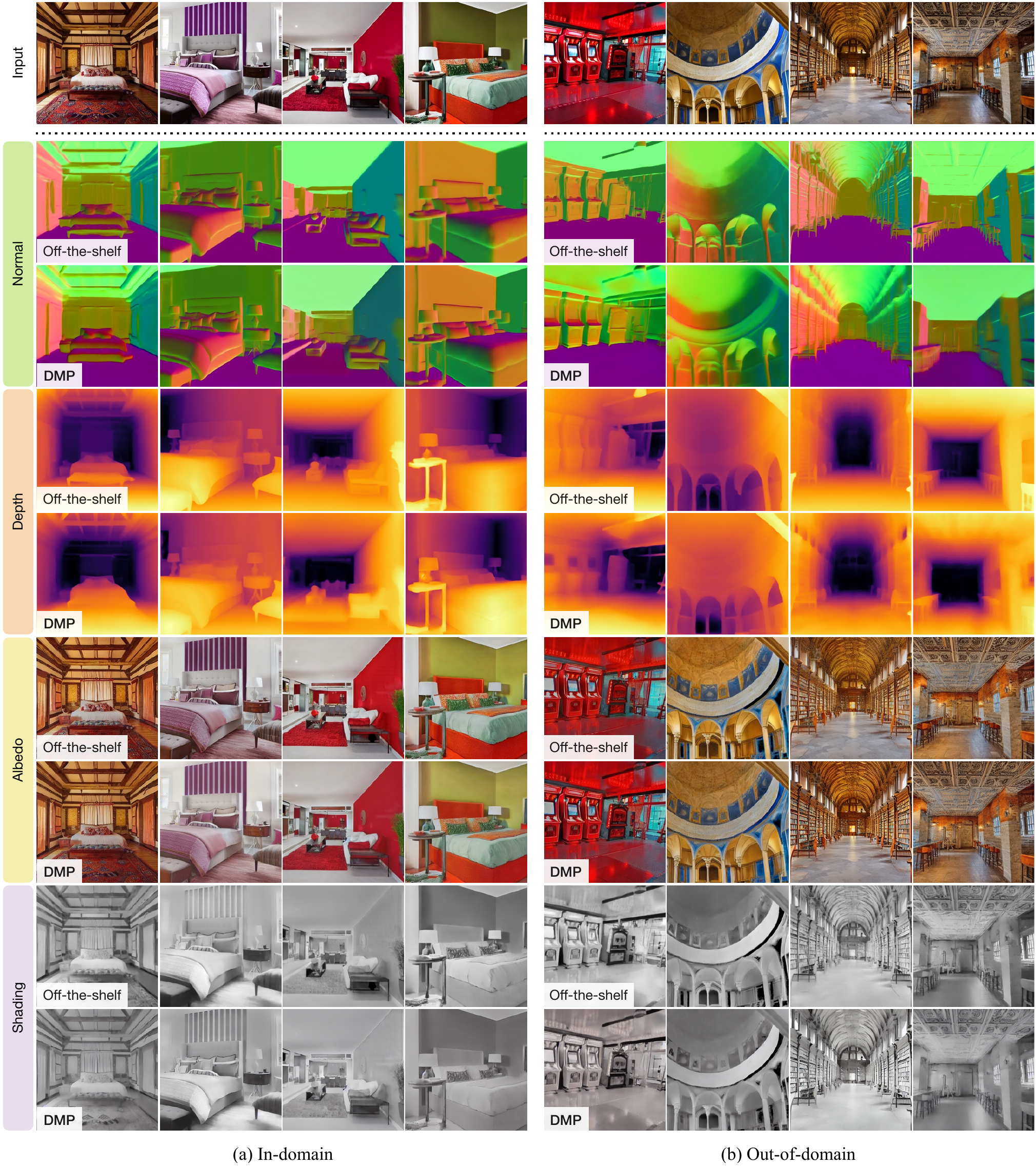}
    \vspace{-2mm}
    \caption{\textbf{Qualitative results.} The first row shows the input images. In the following, every two rows show the results predicted by the off-the-shelf predictors (which we considered as pseudo ground truth) and those by the proposed method.}
    \vspace{-2mm}
    \label{fig:result}
\end{figure*}

\Paragraph{Datasets.} 
To better analyze generalizability by varying training and test data domains, we conduct the evaluation with synthetic images.
We first generate diverse text descriptions using a large language model~\cite{muennighoff-etal-2023-crosslingual} by filling a \emph{keyword} in a prompt template modified from the one used in Parmar et al.~\cite{pix2pixzero}.
We then use a text-to-image diffusion model~\cite{Rombach_2022_CVPR} to synthesize the images according to the text descriptions. 
Second, we use the following off-the-shelf predictors to generate the \emph{pseudo} ground truth for each image: Omnidata v2~\cite{Kar_2022_CVPR} for surface normals, ZoeDepth~\cite{zoedepth} for monocular depth, EVA-2~\cite{EVA02} for semantic segmentation, and PIE-Net~\cite{Das_2022_CVPR} for intrinsic image decomposition (albedo, shading). Finally, we follow the protocol used in Bhattad et al.~\cite{bhattad2023stylegan} to generate a set of training data, and three sets of test data:
\begin{compactitem}
\item Training set: We generate $10$K labeled images using the keyword ``bedroom''.
\item In-domain test set: We generate $2$K labeled images using the keyword ``bedroom''.
\item Out-of-domain test set: We use the $409$ category names of the indoor scenes in the SUN dataset~\cite{xiao2016sun} as the keywords to generate $2$K labeled images. 
The set is considered to be out-of-domain compared to the training set. 
Nevertheless, the off-the-shelf models that provide the pseudo ground truth still work well since the images belong to normal indoor scenes.
\item \emph{Arbitrary} test set: We use random textual descriptions to generate the images.
Since the generated images are almost free-form, the off-the-shelf models cannot provide proper predictions.
Therefore, we consider the off-the-shelf approaches as compared methods and present only the visual comparisons.
\end{compactitem}
Note that we present the quantitative results only on the in-domain test set for semantic segmentation due to the semantic label set constraint.
In addition, we show qualitative comparisons of bedroom images using diverse styles (e.g., cyberpunk, comic) to understand the generalization ability.

\begin{table*}
    \centering \footnotesize
    \caption{\textbf{Quantitative comparisons on semantic segmentation.} We compute the metrics using the estimated results and the pseudo ground truth generated by the off-the-shelf predictors.}
    \vspace{-2mm}
    \begin{tabular}{lcccccccccccc}
        \toprule
         & \multicolumn{2}{c}{Bed} & \multicolumn{2}{c}{Pillow} & \multicolumn{2}{c}{Lamp} & \multicolumn{2}{c}{Window} & \multicolumn{2}{c}{Painting} & \multicolumn{2}{c}{Mean} \\
         \cmidrule(lr){2-3} \cmidrule(lr){4-5} \cmidrule(lr){6-7} \cmidrule(lr){8-9} \cmidrule(lr){10-11} \cmidrule(lr){12-13}
         & Acc$\uparrow$ & mIoU$\uparrow$ & Acc$\uparrow$ & mIoU$\uparrow$ & Acc$\uparrow$ & mIoU$\uparrow$ & Acc$\uparrow$ & mIoU$\uparrow$ & Acc$\uparrow$ & mIoU$\uparrow$ & Acc$\uparrow$ & mIoU$\uparrow$ \\
        \midrule
        SPADE~\cite{Park_2019_CVPR} & \underline{0.8677} & \underline{0.6370} & \underline{0.5861} & \underline{0.3473} & \underline{0.3659} & \underline{0.2084} & \underline{0.6925} & \underline{0.5627} & \underline{0.5249} & \underline{0.3826} & \underline{0.6074} & \underline{0.4276} \\
        DRIT++~\cite{DRIT_plus}  & 0.8485 & 0.4587 & 0.2427 & 0.1435 & 0.1218 & 0.0776 & 0.3023 & 0.2414 & 0.2579 & 0.2114 & 0.3546 & 0.2265 \\
        \midrule
        SDEdit~\cite{meng2022sdedit} & 0.0958	& 0.0901 & 0.3824 & 0.0864 & 0.1522 & 0.0651 & 0.4501 & 0.2593 & 0.1333 & 0.0746 & 0.2428 & 0.1151 \\
        DDIB~\cite{su2023dual}   & 0.3984 & 0.3040 & 0.2256 & 0.0637 & 0.1630 & 0.0593 & 0.4741 & 0.2896 & 0.1728 & 0.0881 & 0.2868 & 0.1609 \\
        IP2P (learned)~\cite{Brooks_2023_CVPR}   & 0.0714 & 0.0620 & 0.0086 & 0.0042 & 0.0228 & 0.0116 & 0.3532 & 0.1699 & 0.0386 & 0.0192 & 0.0989 & 0.0534 \\
        VISII~\cite{visii}  & 0.0060 & 0.0059 & 0.0261 & 0.0136 & 0.0014 & 0.0011 & 0.2576 & 0.1772 & 0.0013 & 0.0012 & 0.0585 & 0.0398 \\
        \midrule
        \NAME  & \textbf{0.8947} & \textbf{0.8506} & \textbf{0.5871} & \textbf{0.3645} & \textbf{0.6399} & \textbf{0.4414} & \textbf{0.8338} & \textbf{0.7335} & \textbf{0.7490} & \textbf{0.6735} & \textbf{0.7409} & \textbf{0.6127} \\
        \bottomrule
    \end{tabular}
    \label{tab:seg}
\end{table*}
\begin{figure*}
    \centering
    \includegraphics[width=\linewidth]{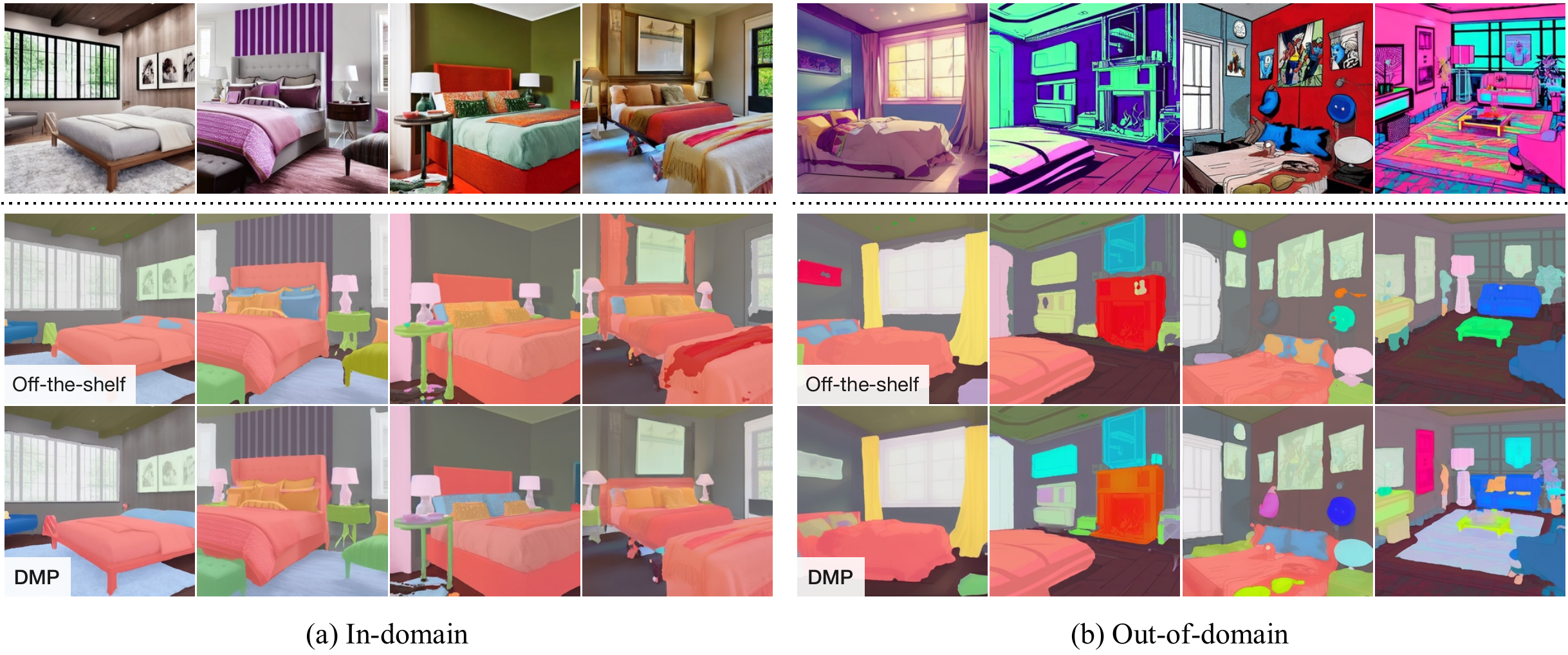}
    \vspace{-7mm}
    \caption{\textbf{Qualitative results on semantic segmentation.} The first, second, and third rows respectively show the input images, pseudo ground truth predicted by an off-the-shelf model, and our results. The out-of-domain samples in (b) are bedroom images in diverse styles.}
    \label{fig:result-seg}
\end{figure*}

\Paragraph{Compared methods.}
We compare our method with the GAN-based image translation methods SPADE~\cite{Park_2019_CVPR} and DRIT++~\cite{mao2019mode, DRIT_plus}.
These models are trained from scratch using the training set (i.e., $10$K labeled bedroom images).
We also include for comparison the following approaches that leverage the pre-trained T2I model as the prior:
\begin{compactitem}
\item \textbf{SDEdit}~\cite{meng2022sdedit}: We fine-tune using the training label images $\{y^i\}^{10\mathrm{K}}_{i=1}$ with the standard diffusion process in Eq. \eqref{eq:diffusion}.
Then we follow the original SDEdit approach that adds the noise to an input $x$ and uses the fine-tuned model to de-noise for generating the output $y$.
\item \textbf{DDIB}~\cite{su2023dual}: We use the same fine-tuned model in SDEdit and adopt the DDIB method to predict the output $y$ from the input $x$.
\item InstructPix2Pix (IP2P)~\cite{Brooks_2023_CVPR}: We evaluate two versions. In \textbf{IP2P (hard)}, we use the pre-defined instructions such as ``make it into the corresponding depth map'' as the input to the model for inference.
For the second version \textbf{IP2P (learned)}, we optimize the token $\ast$ in the input text ``make it into $\ast$'' using the training set.
\item \textbf{VISII}~\cite{visii}: We use their approach for fine-tuning with the training set and inference.
\end{compactitem}

\subsection{3D Property Estimation}

Surface normals~\cite{Dai_2017_CVPR, Eftekhar_2021_ICCV} and depth~\cite{nyu, sunrgbd} are crucial to 3D visual applications such as 3D reconstruction~\cite{Zamir_2018_CVPR} and autonomous driving~\cite{kitti_depth}.
To evaluate normal prediction, we use the average $\mathrm{L1}$ distance and average angular error $\mathrm{Ang}$. For monocular depth, given the ground truth depth $y^i$ and predicted depth $\hat{y}^i$, we use the average relative error $\mathrm{REL}=\frac{1}{M} \sum_{i=1}^M |y^i-\hat{y}^i|/y^i$, percentage of pixels $\delta$ where $\max(y^i/\hat{y}^i, \hat{y}^i/y^i) < 1.25$, and the root mean square error $\mathrm{RMSE}$ of the relative depth.
Specifically, we normalize the ground-truth and predicted depth maps separately to be in the range of $[0,1]$ as the relative depth.

The quantitative comparisons are presented in Table~\ref{tab:normal-depth}, and qualitative results are shown in Figure~\ref{fig:result}.
The proposed approach performs favorably against the compared algorithms in terms of accuracy and generalization ability.
Although the $\mathrm{REL}$ and $\delta$ scores reported for our method degrade from the in-domain to out-of-domain test sets, the $\mathrm{RMSE}$ score of the relative depth remains the same.
This is due to the scene scale difference between the training images (bedroom) and test images (larger spaces such as sports fields).
The $\mathrm{RMSE}$ measurement is computed based on relative depth, which is more resistant to the scene scale change.
Consistent with the quantitative results, we observe that the depth predicted by our method is in the correct order for the out-of-domain images in Figure~\ref{fig:result}.

\Paragraph{Generalization ability.}
We demonstrate the generalizability of the proposed approach in Figure~\ref{fig:superood}, where we use \emph{arbitrary} images as the inputs.
Although our method is fine-tuned with only $10$K bedroom images with labels, it faithfully estimates the 3D property even on those images that the off-the-shelf methods fail to handle.

\begin{table}
    \centering \footnotesize
    \vspace{-2mm}
    \caption{\textbf{Quantitative comparisons on intrinsic image decomposition.} We compute the metrics to measure the difference between the estimated results and the pseudo ground truth created by the off-the-shelf predictors.
    }
    \vspace{-2mm}
    \begin{tabular}{lcccc} 
        \toprule
        & \multicolumn{2}{c}{Albedo} & \multicolumn{2}{c}{Shading} \\
        \cmidrule(lr){2-3} \cmidrule(lr){4-5}
        & In & Out & In & Out \\
        \midrule
         SPADE \cite{Park_2019_CVPR} & \textbf{0.0021} & \textbf{0.0030} & \textbf{0.0031} & \textbf{0.0040} \\
         DRIT++ \cite{DRIT_plus}    & 0.0296 & 0.0392 & 0.0309 & 0.0408 \\
         \midrule
         SDEdit \cite{meng2022sdedit}   & 0.0375 & 0.0471 & 0.0501 & 0.0671 \\
         DDIB \cite{su2023dual}         & 0.0411 & 0.0443 & 0.0403 & 0.0557 \\
         IP2P (hard) \cite{Brooks_2023_CVPR}    & 0.0329 & 0.0479  & 0.0361& 0.0421 \\
         IP2P (learned) \cite{Brooks_2023_CVPR} & 0.0215 & 0.0250 & 0.0290 & 0.0309 \\
         VISII \cite{visii}             & 0.0145 & 0.0246 & 0.0275 & 0.0285 \\
         \midrule
         \NAME   & \underline{0.0041} & \underline{0.0064} & \underline{0.0051} & \underline{0.0070} \\
        \bottomrule
    \end{tabular}
    \label{tab:albedo-shading}
\end{table}
\begin{table} 
    \caption{\textbf{Effect of different parameterizations.} We fine-tune the U-Net model to predict different signals in each reverse diffusion step to get the final outputs: predicting the input image $x$, predicting the output $y$, and v-prediction described in Eq. \eqref{eq:loss}. The experiment is conducted on the surface normal prediction task.
    }
    \centering \footnotesize
    \begin{tabular}{lcccc}
        \toprule
        & \multicolumn{2}{c}{In-domain} & \multicolumn{2}{c}{Out-of-domain} \\
        \cmidrule(lr){2-3} \cmidrule(lr){4-5}
        & L1$\downarrow$ & Ang$\downarrow$ & L1$\downarrow$ & Ang$\downarrow$ \\
        \midrule
        Predicting $x$  & 0.0736 & 0.1629 & 0.1319 & 0.2764 \\
        Predicting $y$  & 0.0590 & 0.1291 & 0.0888 & 0.1914 \\
        \midrule
        v-prediction & \textbf{0.0514} & \textbf{0.1156} & \textbf{0.0872} & \textbf{0.1886} \\
        \bottomrule
    \end{tabular}
    \label{tab:ablate-pred}
\end{table}
\begin{figure}
    \centering
    \includegraphics[width=\linewidth]{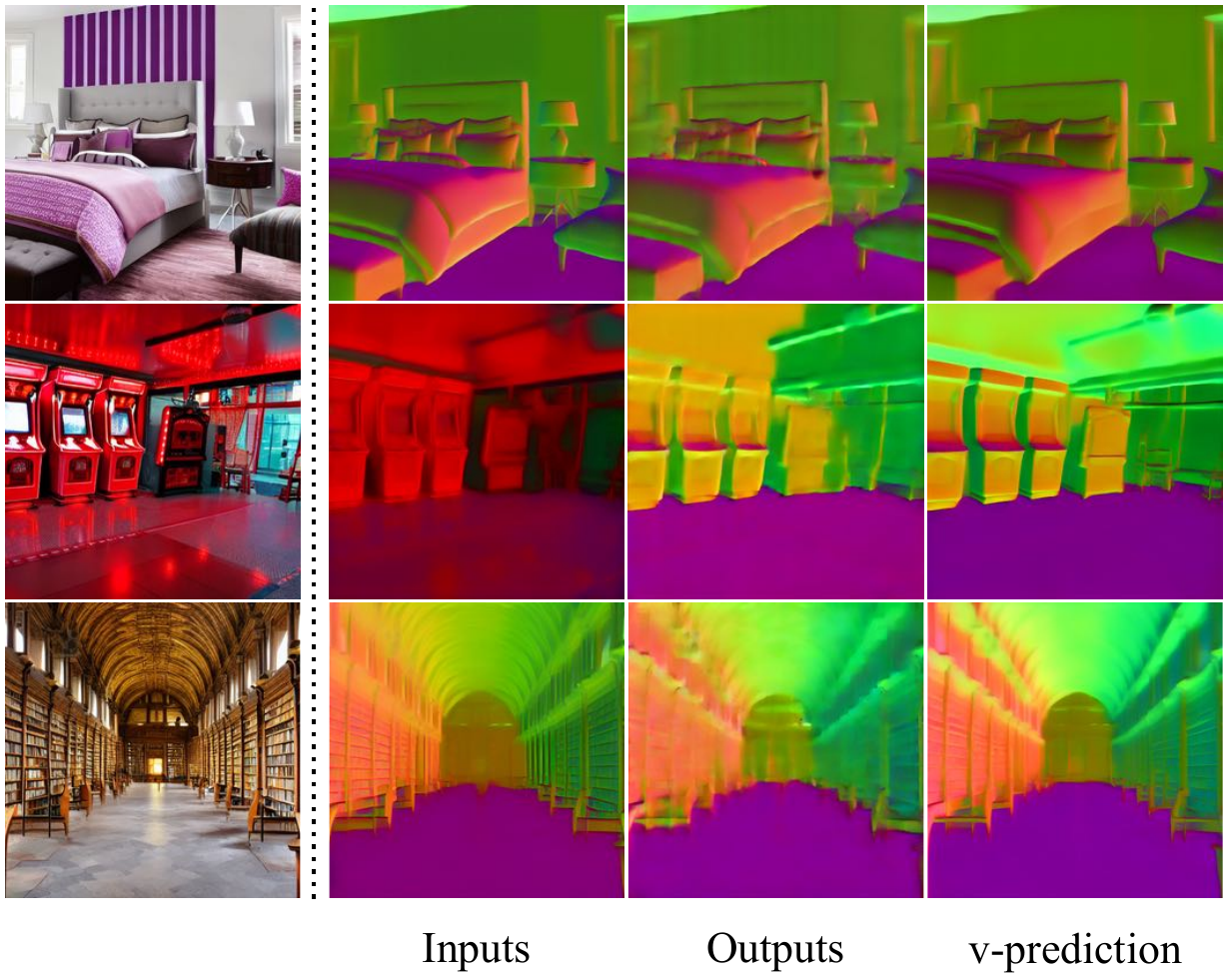}
    \vspace{-5mm}
    \caption{
    \textbf{Qualitative comparisons between various parameterizations.} We fine-tune the U-Net model to predict different signals in each reverse diffusion step to get the final output: predicting the input image $x$, predicting the output $y$, and v-prediction described in Eq. \eqref{eq:loss}. 
    }
    \label{fig:ablate-pred}
\end{figure}
\subsection{Semantic Segmentation}

Semantic segmentation~\cite{ade20k} is a fundamental visual understanding task. 
Since the prediction is categorical, we use a simple conversion for regression models to predict discrete labels. 
We first generate ground-truth labels using the off-the-shelf EVA-2~\cite{EVA02} model. 
The label maps are then transformed into color maps where each category has different colors. 
The training and inference of the diffusion model are conducted using the color maps (in the RGB space). 
During inference, the predicted color maps are converted to categorical label maps by assigning each pixel to its nearest category in the color space.

We report the intersection over union (IoU) and accuracy to measure the in-domain performance in Table~\ref{tab:seg}.
The prediction by the proposed scheme is reasonable across all categories compared to the existing methods.
Figure~\ref{fig:result-seg} demonstrates the results of our method and pseudo ground truth with the in- and out-of-domain (i.e., bedroom images in diverse styles) input images. 
Particularly in out-of-domain examples, our model gives favorable predictions compared to the off-the-shelf approach, e.g., the painting of the first, the curtain of the second, the window of the third, and the carpet of the fourth images in Figure~\ref{fig:result-seg} (b).
This validates our idea that leverages the pre-trained T2I diffusion model as the prior for better generalization.

\subsection{Intrinsic Image Decomposition}

Intrinsic image decomposition~\cite{bell_2014_intrinsic} recovers albedo and shading properties from RGB images. 
It facilitates applications such as recoloring~\cite{beigpour_iccv_2011} and relighting~\cite{Shu_2017_CVPR}. 
Similar to PIE-Net~\cite{Das_2022_CVPR}, we use the mean square error $\mathrm{MSE}$ as the evaluation metric to compute the distance between the predicted and pseudo ground-truth estimations.
We showcase the qualitative results in Figure~\ref{fig:result} and quantitative comparisons in Table~\ref{tab:albedo-shading}.
The reported errors of most methods are an order of magnitude larger than ours.
Notably, while our method and SPADE work well in both in- and out-of-domain settings, we found our method less influenced by artifacts in pseudo ground truth and to offer more reasonable estimations. More details are provided in the supplementary document.

\begin{figure}
    \centering
    \includegraphics[width=0.5\linewidth]{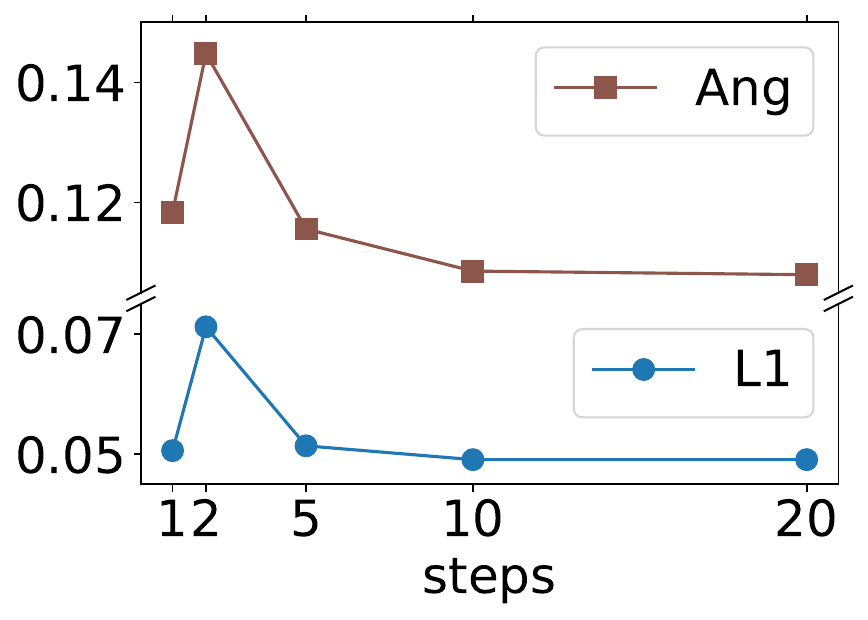}
    \vspace{-3mm}
    \caption{\textbf{Effect of varying the numbers of diffusion steps.} We report the in-domain performance on the surface normal prediction task.
    In the case of single-step, we train the U-Net model to directly predict outputs from input images.
    }
    \label{fig:ablate-step-plot}
\end{figure}
\begin{figure}
    \centering
    \includegraphics[width=0.765\linewidth]{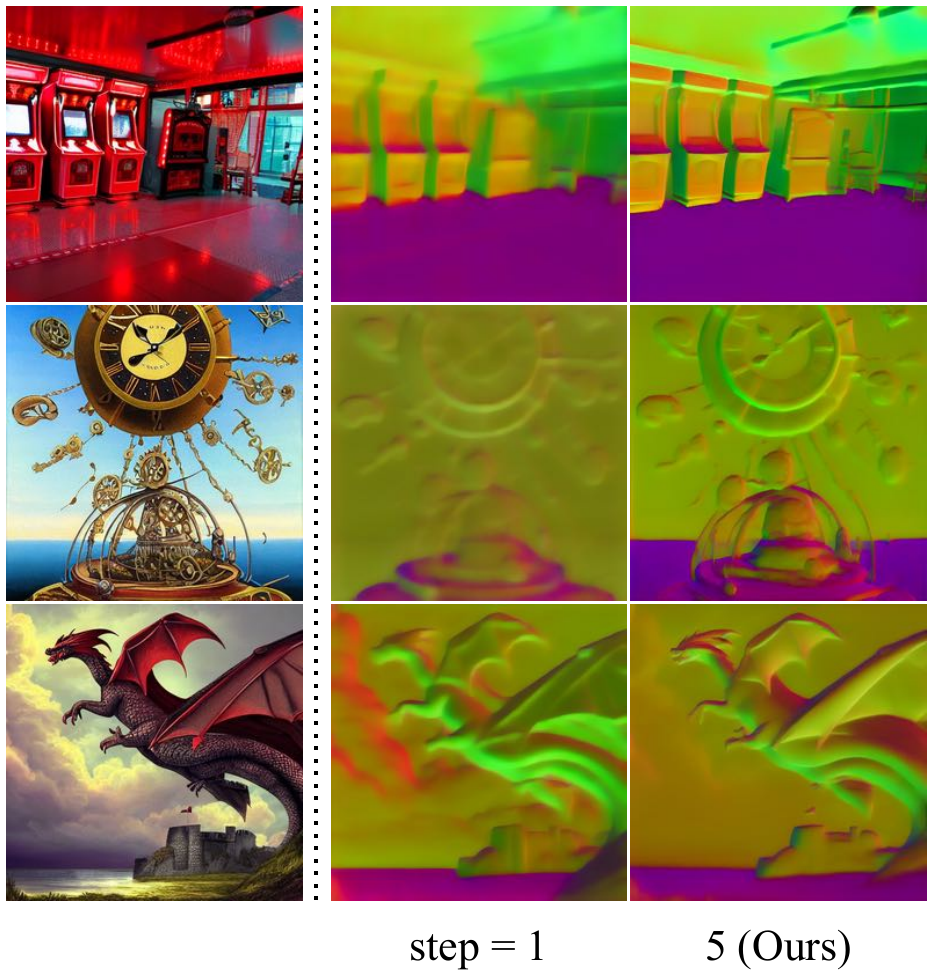}
    \vspace{-1mm}
    \caption{\textbf{Single vs. multiple diffusion steps.} 
    In the single-step approach, we train the U-Net network to directly predict outputs from input images.
    The single-step approach does not generalize to unseen images well.
    }
    \label{fig:ablate-step}
\end{figure}

\subsection{Ablation Study}
\label{sec:exp_ablation}

\Paragraph{Parameterization.}
We study the effect of employing various parameterizations.
Specifically, we fine-tune the U-Net model to make different predictions in each reverse diffusion step for obtaining the final prediction: 1) predicting the input image $x$, predicting the output $y$ (similar to $x_0$-prediction in standard diffusion models), and v-prediction described in Eq. \eqref{eq:loss}.
We formulate each parameterization in detail in the supplementary document.
The quantitative results are shown in Table~\ref{tab:ablate-pred} and the qualitative comparisons are presented in Figure~\ref{fig:ablate-pred}.
Predicting the input image $x$ generates accurate results in the in-domain setting.
However, it fails to generalize to unseen domains.
On the other hand, predicting the output $y$ demonstrates preferred generalization capability, but tends to produce blurry results with few details.
We choose to use the v-prediction approach as it produces accurate results of arbitrary images.

\Paragraph{Number of diffusion steps.}
We analyze the performance of our method with different numbers of diffusion steps in the inference stage.
The quantitative comparisons on the normal prediction are presented in Figure~\ref{fig:ablate-step-plot}.
Using $5$ steps strikes a good balance between inference speed and accuracy.
Furthermore, we study an extreme case of using a single step. We train the U-net to directly predict the output $y$ from the input image $x$.
Although it shows competitive performance in the in-domain setting according to Figure~\ref{fig:ablate-step-plot}, we find that the performance degrades significantly with the \emph{arbitrary} input images, as examples shown in Figure~\ref{fig:ablate-step}.
Considering the generalizability, inference speed, and accuracy, we use $5$ generation steps for all the other experiments.
\section{Conclusion}

In this work, we leverage a pre-trained T2I diffusion model for generalizable dense prediction.
The core of our method is the design of the deterministic diffusion process that adapts the stochastic T2I framework for the deterministic prediction tasks.
With low-rank approximation, the proposed approach learns the target tasks while retaining the inherent generalization ability of the T2I model.
We show that with only a small number of labeled training data in a limited domain (i.e., $10$K bedroom images), our \NAME scheme makes faithful predictions on \emph{arbitrary} images.
We believe that this work establishes a foundation for achieving ultimately-generalizable visual understanding in the future.

\clearpage

{
    \small
    \bibliographystyle{ieeenat_fullname}
    \bibliography{main}
}

\appendix
\clearpage
% \setcounter{page}{1}
% \maketitlesupplementary

\section{Parametrizations}

As described in Section 3.2, we empirically find that parametrizing the U-Net model through estimating v-prediction~\cite{salimans2022progressive} performs favorably against predicting inputs or outputs. We detail the formulation of predicting inputs and outputs as follows. The U-Net model $\hat{x}_\theta$ predicting inputs is fine-tuned with the mean square loss:
\begin{equation}
    L=\mathbb{E}_{(x,y),t}\big[\|x - \hat{x}_\theta(y_t, t)\|^2_2\big],
\end{equation}
and the reverse diffusion process is formulated as
\begin{equation}
    \begin{aligned}
    y_{t-1} &= \sqrt{\bar{\alpha}_{t-1}} \left( \frac{y_t - \sqrt{1 - \bar{\alpha}_t} \hat{x}_\theta(y_t, t)}{\sqrt{\bar{\alpha}_t}} \right)\\
    &+ \sqrt{1 - \bar{\alpha}_{t-1}} \hat{x}_\theta(y_t, t) \hspace{7mm} t=[T,\cdots,1],
    \end{aligned}
\end{equation}
The U-Net model $\hat{y}_\theta$ predicting outputs is optimized with the loss function:
\begin{equation}
    L=\mathbb{E}_{(x,y),t}\big[\|y - \hat{y}_\theta(y_t, t)\|^2_2\big],
\end{equation}
and the reverse diffusion process is
\begin{equation}
    \begin{aligned}
    y_{t-1}&=\sqrt{\bar{\alpha}_{t-1}}\hat{y}_\theta(y_t, t) + \sqrt{1 - \bar{\alpha}_{t-1}}x \\
    t&=[T,\cdots,1],
    \end{aligned}
\end{equation}

\section{Additional Experimental Results}

\subsection{Reliability of Off-the-Shelf Estimators}

\begin{figure}
    \centering
    \includegraphics[width=.9\linewidth]{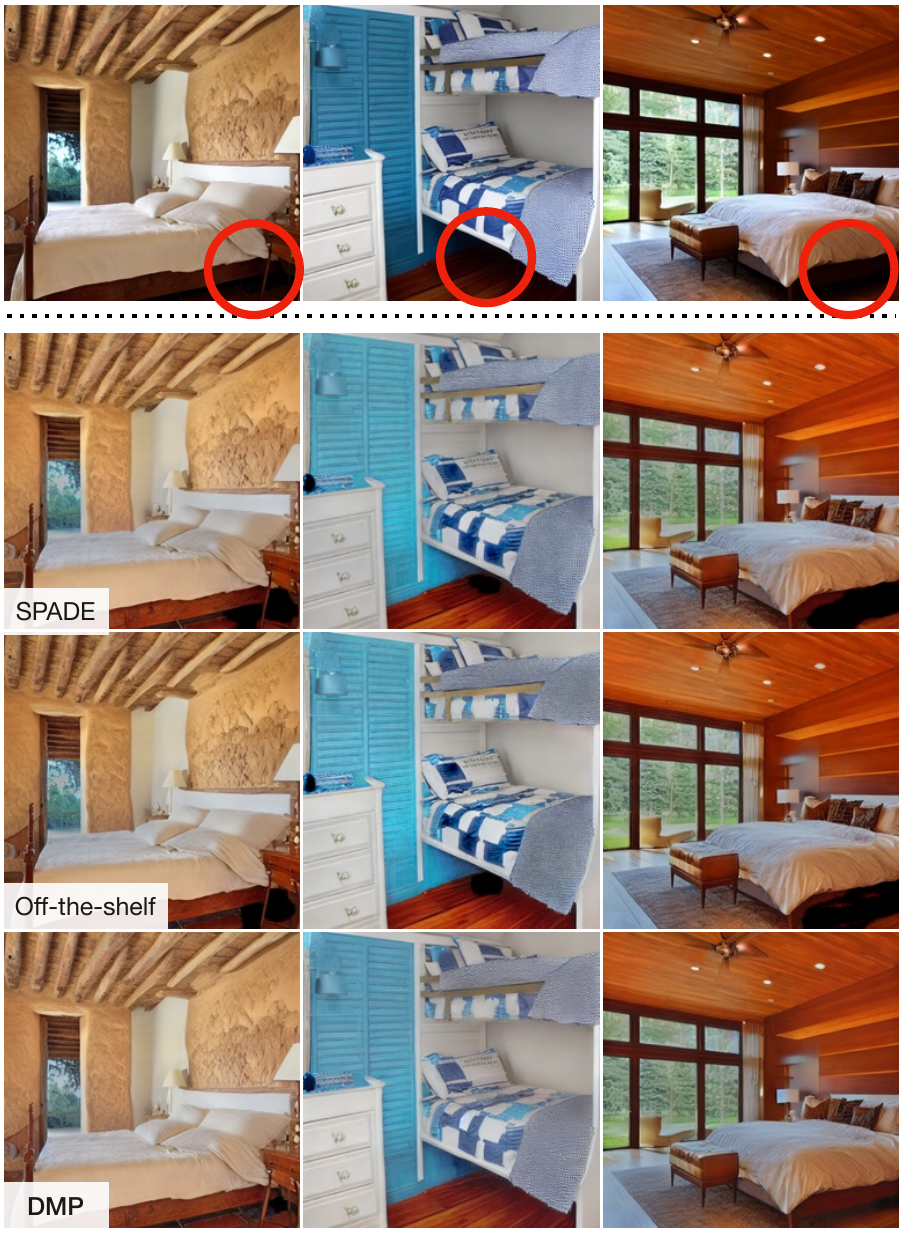}
    \caption{\textbf{Qualitative comparisons on albedo estimation}. SPADE~\cite{park2019SPADE} and the off-the-shelf approaches generate artifacts in dark areas.}
    \label{fig:ref-issue}
\end{figure}

We indicate that the off-the-shelf estimators are not always reliable, especially the approach for intrinsic image decomposition. We demonstrate with the example of albedo estimation in \cref{fig:ref-issue} that the off-the-shelf approach generates apparent artifacts in shadow areas, such as corners or floors under beds. The approach fails to recover the correct albedo but instead generates black patches. Consequently, SPADE also learns this pattern, but our model tends to correct artifacts by performing accurate estimation, manifesting the ability of generalization.

\subsection{Real-World Evaluation}

\paragraph{NYU Depth v2.} Following \citet{ke2023repurposing}, we evaluate our method on NYU Depth v2~\cite{nyu} according to the protocol of affine-invariant depth evaluation~\cite{midas}. We generate prompts with BLIP-2~\cite{blip2} to use the model trained on synthetic bedroom images. We scale and shift the depth predictions to align ground truths by solving least-square fitting. The comparison against other approaches is shown in \cref{tab:nyu}. \NAME performs comparably with some previous methods trained with large-scale data.
\begin{table}
    \centering \footnotesize
    \caption{Comparison of performance on NYU Depth v2~\cite{nyu}.}
    \begin{tabular}{lrrcc}
        \toprule
         & \multicolumn{2}{c}{\# Training Samples} & \multicolumn{2}{c}{NYU v2} \\
         \cmidrule(lr){2-3} \cmidrule(lr){4-5}
         & Real    & Synthetic & REL$\downarrow$ & $\delta\uparrow$ \\
        \midrule
        MiDaS~\cite{midas}                  & 2M    & --    & 11.1 & 88.5 \\
        Omnidata~\cite{Eftekhar_2021_ICCV}  & 11.9M & 310K  & 7.4  & 94.5 \\
        DPT~\cite{Ranftl_2021_ICCV}         & 1.2M  & 188K  & 9.8  & 90.3 \\
        Painter~\cite{Wang_2023_CVPR}       & 24K   & --    & 8.0  & 95.0 \\
        Marigold~\cite{ke2023repurposing}   & --    & 74K   & 5.5  & 96.4 \\ 
        \midrule
        \NAME    & --    & 10K   & 12.0 & 86.5 \\
        \bottomrule
    \end{tabular}
    \label{tab:nyu}
\end{table}

\paragraph{ADE20K.} We also investigate the performance of semantic segmentation with more classes, \eg 150 classes in ADE20K~\cite{ade20k}. We follow the encoding strategy proposed by \citet{Wang_2023_CVPR} and convert class indices into 3-digit numbers with a $b$-base system, which can be represented in the RGB space. However, the performance is unsatisfying (lower than $20\%$ accuracy). With the number of classes increasing, the differences between colors are less distinguishable. The unlabeled areas, which can be simply ignored when calculating loss in the image space, become hard to tackle in the latent space. We leave the application of real-world many-class semantic segmentation for future exploration.

We conduct another analysis with a subset of ADE20K containing only images with beds. The train-test split is constructed by applying the same filtering to the original splits, resulting in 1825 training images and 189 test images. We also generate prompts with BLIP-2. The results are presented in \cref{tab:ade20k}. The performance across real and synthetic domains is similar, especially for large items.
\begin{table}[t]
    \centering \footnotesize
    \setlength{\tabcolsep}{3.5pt}
    \caption{Comparison of performance between the subset of ADE20K~\cite{ade20k} and the synthesized bedrooms.}
    \begin{tabular}{cccccccc}
        \toprule
             & & Bed & Pillow & Lamp & Window & Painting & Mean \\
         \midrule
        \multirow{2}{*}{ADE20K} & Acc$\uparrow$ & 0.88 & 0.36 & 0.57 & 0.76 & 0.74 & 0.66 \\
        & mIoU$\uparrow$ & 0.82 & 0.25 & 0.39 & 0.60 & 0.60 & 0.53 \\
        \midrule
        \multirow{2}{*}{Bedrooms} & Acc$\uparrow$ & 0.89 & 0.59 & 0.64 & 0.83 & 0.75 & 0.75 \\
        & mIoU$\uparrow$ & 0.85 & 0.36 & 0.44 & 0.73 & 0.67 & 0.61 \\
        \bottomrule
    \end{tabular}
    \label{tab:ade20k}
\end{table}

\subsection{Additional Ablation Study}

\paragraph{Modeling.}
We vary the trainable layers and the presence of text conditions when fine-tuning the model. Since the example tasks we choose in this work are not directly conditional on text, providing text descriptions might not be necessary. Accordingly, training cross-attention layers is optional. \cref{tab:ablate-cond} shows that text condition and training cross-attention layers help improve the performance of in-domain samples, but the difference in out-of-domain samples is unnoticeable between the settings. This result suggests that we can adopt curated real ground truth datasets without text descriptions for training at the expense of a subtle performance drop. Alternatively, we can generate prompts with image captioning~\cite{blip2}, which may lead to better performance.

\begin{table}
    \caption{\textbf{Analysis of training cross-attention layers and providing text condition.} Both improve the performance of in-domain samples but make little difference in out-of-domain data.}
    \centering \footnotesize
    \begin{tabular}{lcccc}
        \toprule
        & \multicolumn{2}{c}{In-domain} & \multicolumn{2}{c}{Out-of-domain} \\
        \cmidrule(lr){2-3} \cmidrule(lr){4-5}
        & L1$\downarrow$ & Ang$\downarrow$ & L1$\downarrow$ & Ang$\downarrow$ \\
        \midrule
        Self-attn  & 0.0606 & 0.1290 & 0.0890 & 0.1871 \\
        Self-attn + text & 0.0605 & 0.1293 & 0.0876 & \textbf{0.1844} \\
        \midrule
        All attn + text & \textbf{0.0514} & \textbf{0.1156} & \textbf{0.0872} & 0.1886 \\
        \bottomrule
    \end{tabular}
    \label{tab:ablate-cond}
\end{table}

\begin{figure}
    \centering
    \includegraphics[width=0.8\linewidth]{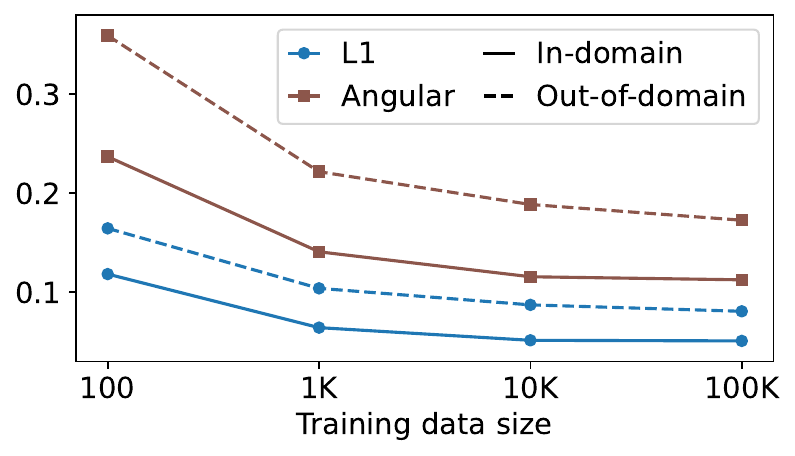}
    \caption{Quantitative performance of normal estimation with different sizes of training data.}
    \label{fig:ablate-size}
\end{figure}

\paragraph{Size of Training Data.} We analyze the effect of varying the size of training data. We compare fine-tuning the model for normal estimation with $100$, $1$K, $10$K, and $100$K generated bedroom images. As shown in \cref{fig:ablate-size}, increasing data size over $10$K improves little performance, so we conduct the other experiments with $10$K training images.

\begin{table}[t]
    \centering \footnotesize
    \caption{NYU Depth v2~\cite{nyu} performance comparison of models trained with real and pseudo ground truths.}
    \begin{tabular}{llcc}
        \toprule
        Dataset & Ground Truth & REL$\downarrow$ & $\delta\uparrow$ \\
        \midrule
        Hypersim~\cite{hypersim} & Real & 13.0 & 85.0 \\
        \midrule
        Bedrooms & Pseudo & 12.0 & 86.5 \\
        \bottomrule
    \end{tabular}
    \label{tab:ablate-quality}
\end{table}
\paragraph{Quality of Training Data.} We examine the influence of data quality by comparing the models trained with real and pseudo ground truth. We use Hypersim~\cite{hypersim} as the real ground truth and evaluate the models with NYU Depth v2~\cite{nyu}. \cref{tab:ablate-quality} shows that there is no significant difference between the two models. The model trained with pseudo ground truth even performs slightly better. We speculate that the data diversity may be an important factor. While Hypersim contains more than 70K images, the images are collected from only 461 scenes. Many of them are variations of camera views and distances. In contrast, the synthetic images, while all of them are bedrooms, are all distinct scenes, which present diverse compositions of objects.

\paragraph{Blending Inputs and Outputs.} IADB~\cite{alphablend} proposes a deterministic framework where the diffusion process is formulated as a series of interpolations between observations and noise. Although their training strategy produces deterministic mapping of observations and noise, the correlation between observation and noise in each pair is stochastic due to unpaired sampling during training. We analyze the applicability of this framework to deterministic dense prediction problems by sampling paired inputs and outputs and fine-tuning from a pre-trained T2I diffusion model. With such adaptation, the differences between their framework and our approach are only the variance schedule and parametrization, where the importance weight of inputs linearly rises through the diffusion process, and the U-Net predicts $y-x$.

\begin{figure}
    \centering
    \includegraphics[width=\linewidth]{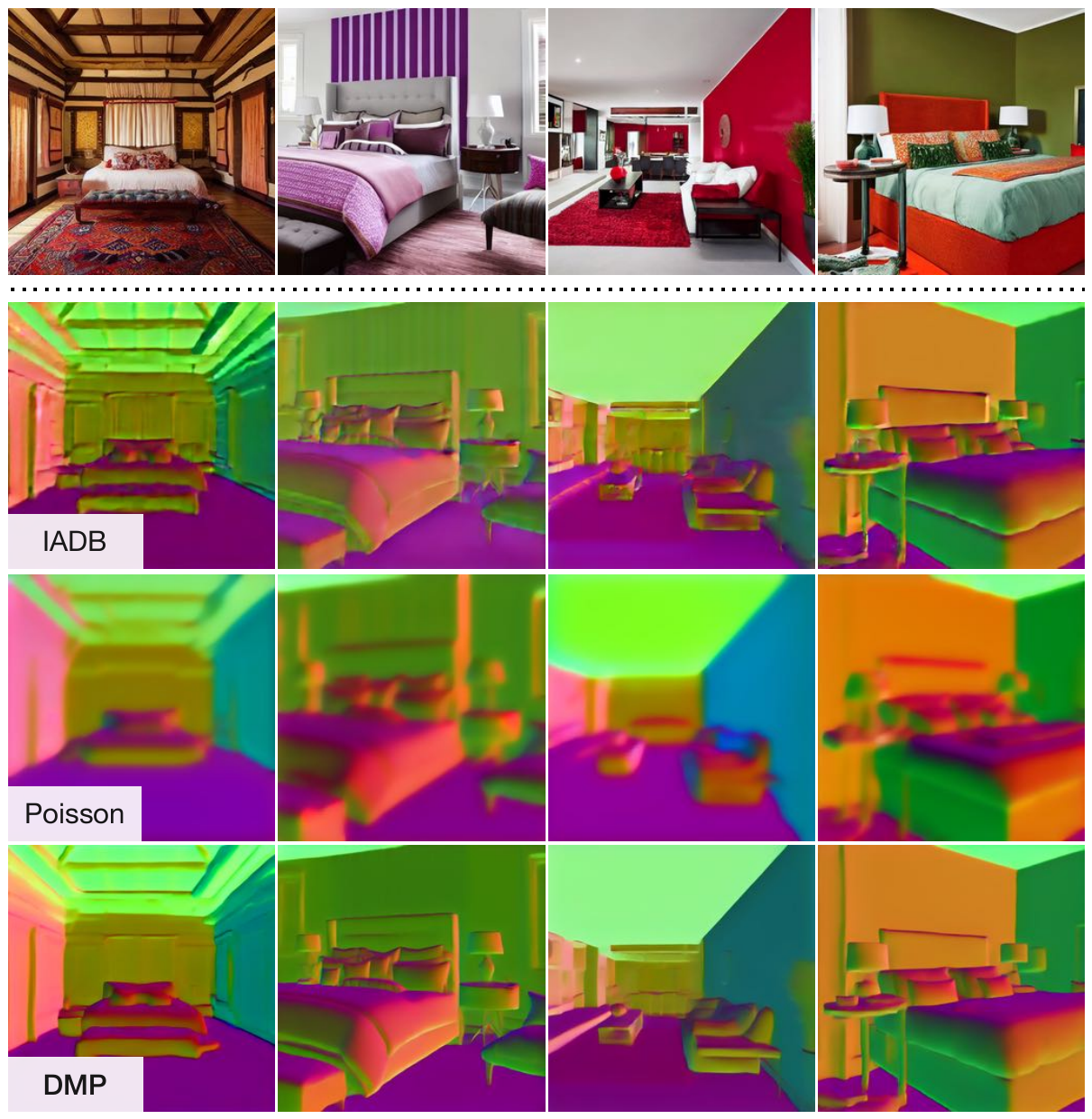}
    \caption{Qualitative comparisons between different blending frameworks.}
    \label{fig:blending}
\end{figure}

\cref{tab:blend-normal} shows the comparison between \NAME and IADB on surface normal estimation, and \cref{tab:blend-depth} is the result of depth estimation. \cref{fig:blending} demonstrates that the images generated by the model fine-tuned through IADB have noise and inaccurate predictions.

In addition to $\alpha$-blending used by \NAME and IADB, we investigate the effect of an advanced blending strategy, Poission blending~\cite{poission}, which blends source and target images by solving a least-square fitting while reserving the gradient of source images. We assume image gradients are meaningful in the latent space. The diffusion process is viewed as increasing the intensity of the mask for selection editing. We adopt an off-the-shelf PyTorch implementation~\cite{Baugh_PIE-torch}. The performance on surface normal estimation is shown in \cref{tab:blend-normal}, and the example outputs in \cref{fig:blending} show that the image quality is unsatisfying.

\begin{table}
    \caption{Comparions with IADB~\cite{alphablend} and Poission blending~\cite{poission} on surface normal estimation.}
    \centering \footnotesize
    \begin{tabular}{lcccc}
         \toprule
        & \multicolumn{2}{c}{In-domain} & \multicolumn{2}{c}{Out-of-domain} \\
        \cmidrule(lr){2-3} \cmidrule(lr){4-5}
        & L1$\downarrow$ & Ang$\downarrow$ & L1$\downarrow$ & Ang$\downarrow$ \\
        \midrule
        IADB~\cite{alphablend} & 0.0675 & 0.1416 & 0.0974 & 0.2017 \\
        Poission~\cite{poission} & 0.0868 & 0.1888 & 0.1201 & 0.2623 \\
        \midrule
        \NAME & \textbf{0.0514} & \textbf{0.1156} & \textbf{0.0872} & \textbf{0.1886} \\
        \bottomrule
    \end{tabular}
    \label{tab:blend-normal}
\end{table}

\begin{table}
    \caption{Comparions with IADB~\cite{alphablend} on depth estimation.}
    \centering \footnotesize
    \setlength{\tabcolsep}{4pt}
    \begin{tabular}{lcccccc}
         \toprule
        & \multicolumn{3}{c}{In-domain} & \multicolumn{3}{c}{Out-of-domain} \\
        \cmidrule(lr){2-4} \cmidrule(lr){5-7}
        & REL$\downarrow$ & $\delta\uparrow$ & RMSE$\downarrow$ & REL$\downarrow$ & $\delta\uparrow$ & RMSE$\downarrow$ \\
        \midrule
        IADB~\cite{alphablend} & 0.3099 & 0.4982 & 0.1165 & 0.5049 & 0.3132 & 0.1467 \\
        \midrule
        \NAME & \textbf{0.1072} & \textbf{0.8861} & \textbf{0.1020} & \textbf{0.2117} & \textbf{0.6395} & \textbf{0.1360} \\
        \bottomrule
    \end{tabular}
    \label{tab:blend-depth}
\end{table}

\subsection{Additional Comparison}

\begin{table*}[t]
    \centering \footnotesize
    \caption{Quantitative comparisons with ControlNet~\cite{Zhang_2023_ICCV} and Palette~\cite{palette} on 3D property estimation and intrinsic image decomposition.}
    \addtolength{\tabcolsep}{-0.6mm}
    \begin{tabular}{lcccccccccccccc}
        \toprule
         & \multicolumn{4}{c}{Normal} & \multicolumn{6}{c}{Depth} & \multicolumn{2}{c}{Albedo} & \multicolumn{2}{c}{Shading} \\
         \cmidrule(lr){2-5} \cmidrule(lr){6-11} \cmidrule(lr){12-13} \cmidrule(lr){14-15}
         & \multicolumn{2}{c}{In} & \multicolumn{2}{c}{Out} & \multicolumn{3}{c}{In} & \multicolumn{3}{c}{Out} & In & Out & In & Out \\
         \cmidrule(lr){2-3} \cmidrule(lr){4-5} \cmidrule(lr){6-8} \cmidrule(lr){9-11} \cmidrule(lr){12-12} \cmidrule(lr){13-13} \cmidrule(lr){14-14} \cmidrule(lr){15-15}
         & L1$\downarrow$ & Ang$\downarrow$ & L1$\downarrow$ & Ang$\downarrow$ & REL$\downarrow$ & $\delta\uparrow$ & RMSE$\downarrow$ & REL$\downarrow$ & $\delta\uparrow$ & RMSE$\downarrow$ & MSE$\downarrow$ & MSE$\downarrow$ & MSE$\downarrow$ & MSE$\downarrow$ \\
        \midrule
         ControlNet~\cite{Zhang_2023_ICCV} & 0.1021 & 0.2216 & 0.1862 & 0.4032 & 0.1739 & 0.7681 & 0.1287 & 0.4398 & 0.4004 & 0.2253 & 0.0302 & 0.0402 & 0.0265 & 0.0336 \\
         Palette~\cite{palette} & 0.1643 & 0.3642 & 0.1881 & 0.4160 & 0.6889 & 0.2626 & 0.3604 & 1.0535 & 0.2270 & 0.4203 & 0.0203 & 0.0199 & 0.0304 & 0.0260 \\
         \midrule
         \NAME    & \textbf{0.0514} & \textbf{0.1156} & \textbf{0.0872} & \textbf{0.1886} & \textbf{0.1072} & \textbf{0.8861} & \textbf{0.0041} & \textbf{0.2117} & \textbf{0.6395} & \textbf{0.1360} & \textbf{0.0051} & \textbf{0.0064} & \textbf{0.1020} & \textbf{0.0070} \\
        \bottomrule
    \end{tabular}
    \label{tab:ctrl-palette}
\end{table*}

\begin{table*}[t]
    \centering \footnotesize
    \caption{Quantitative comparisons with ControlNet~\cite{Zhang_2023_ICCV} and Palette~\cite{palette} on semantic segmentation.}
    \begin{tabular}{lcccccccccccc}
        \toprule
         & \multicolumn{2}{c}{Bed} & \multicolumn{2}{c}{Pillow} & \multicolumn{2}{c}{Lamp} & \multicolumn{2}{c}{Window} & \multicolumn{2}{c}{Painting} & \multicolumn{2}{c}{Mean} \\
         \cmidrule(lr){2-3} \cmidrule(lr){4-5} \cmidrule(lr){6-7} \cmidrule(lr){8-9} \cmidrule(lr){10-11} \cmidrule(lr){12-13}
         & Acc$\uparrow$ & mIoU$\uparrow$ & Acc$\uparrow$ & mIoU$\uparrow$ & Acc$\uparrow$ & mIoU$\uparrow$ & Acc$\uparrow$ & mIoU$\uparrow$ & Acc$\uparrow$ & mIoU$\uparrow$ & Acc$\uparrow$ & mIoU$\uparrow$ \\
        \midrule
        ControlNet~\cite{Zhang_2023_ICCV} & 0.5215 & 0.4820 & 0.3540 & 0.1436 & 0.4275 & 0.2936 & 0.4999 & 0.4190 & 0.3823 & 0.3257 & 0.4370 & 0.3328 \\
        Palette~\cite{palette} & 0.0347 & 0.0329 & 0.0019 & 0.0018 & 0.0013 & 0.0012 & 0.0119 & 0.0119 & 0.0005 & 0.0005 & 0.0101 & 0.0097 \\
        \midrule
        \NAME  & \textbf{0.8947} & \textbf{0.8506} & \textbf{0.5871} & \textbf{0.3645} & \textbf{0.6399} & \textbf{0.4414} & \textbf{0.8338} & \textbf{0.7335} & \textbf{0.7490} & \textbf{0.6735} & \textbf{0.7409} & \textbf{0.6127} \\
        \bottomrule
    \end{tabular}
    \label{tab:ctrl-palette-seg}
\end{table*}
\paragraph{ControlNet.}
ControlNet~\cite{Zhang_2023_ICCV} proposes a conditional text-to-image framework with additional control, such as edges or human poses, which constrains structures and layouts of output images. Since it is also an image-to-image generative model, we train it to take input images as control and output estimations. The performance of estimating 3D properties and intrinsic images is presented in \cref{tab:ctrl-palette}, and the segmentation results are shown in \cref{tab:ctrl-palette-seg}. It demonstrates weaker generalizability than our approach.

\begin{figure}
    \centering
    \includegraphics[width=\linewidth]{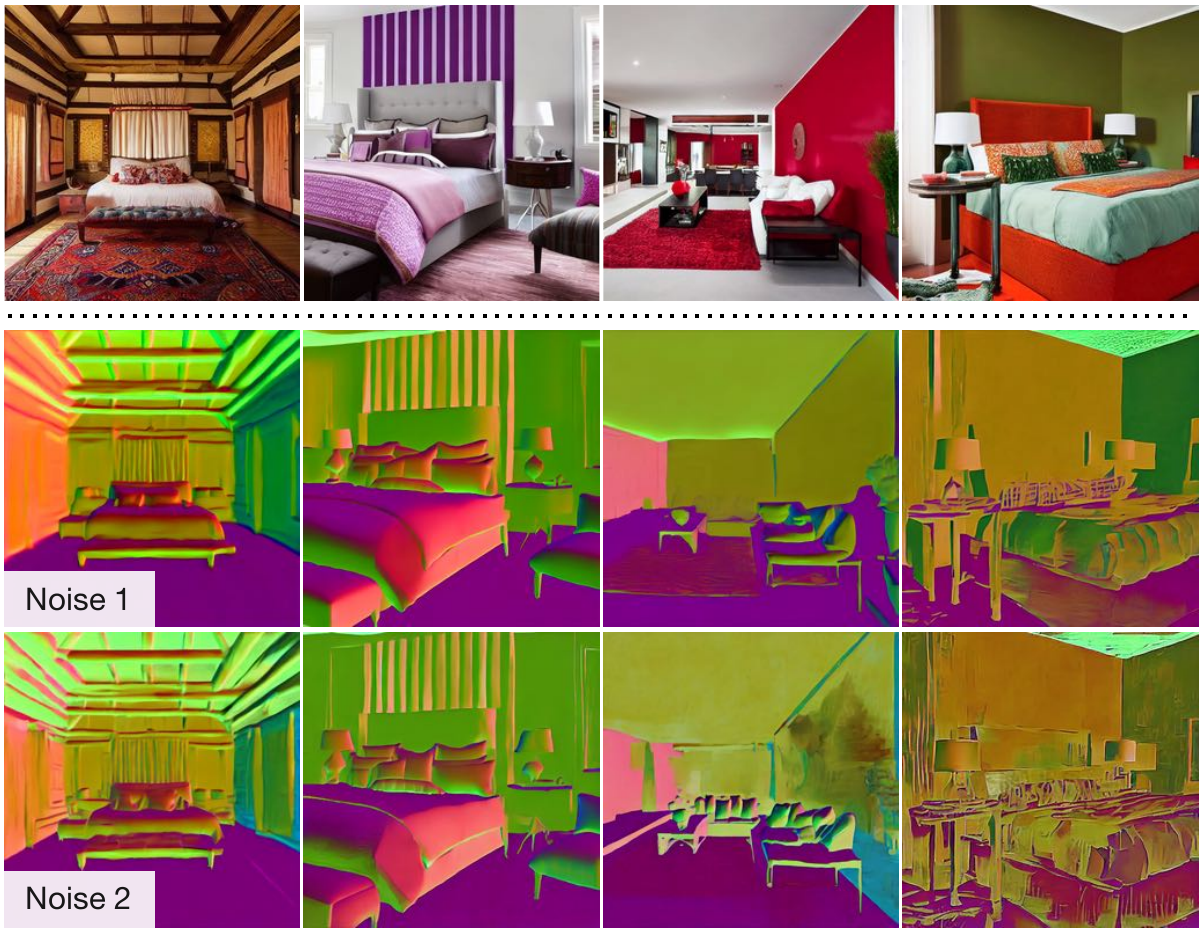}
    \caption{\textbf{Results of ControlNet with different initial noise.} The outputs are not deterministic.}
    \label{fig:controlnet}
\end{figure}

In addition, we analyze the influence of varying initial noise in \cref{fig:controlnet}. While the rough structures of the images are controlled by the input images, the initial noise alters the details of estimations. This variation is not tolerated for dense prediction.

\paragraph{Palette.}
Besides training GAN-based generative models from scratch and fine-tuning pre-trained diffusion models with the approaches listed in Section 4.1, we additionally include training an image-to-image diffusion model from scratch for comparison. Following the design of Palette~\cite{palette}, we expand the input layers of the U-Net to encode the concatenation of input and output images, with the U-Net parameterized to predict noise. The same autoencoder in the pre-trained diffusion model is also adopted. The performance is shown in \cref{tab:ctrl-palette} and \cref{tab:ctrl-palette-seg}, which indicates the inability of this approach to handle categorical label maps.

\subsection{Improving Compared Methods}

Since the results of GAN-based generative models consistently outperform diffusion-based models in our experiments, we seek performance enhancement for diffusion-based approaches. All experiments are conducted on in-domain surface normal estimation. 

\begin{figure}
    \centering
    \includegraphics[width=\linewidth]{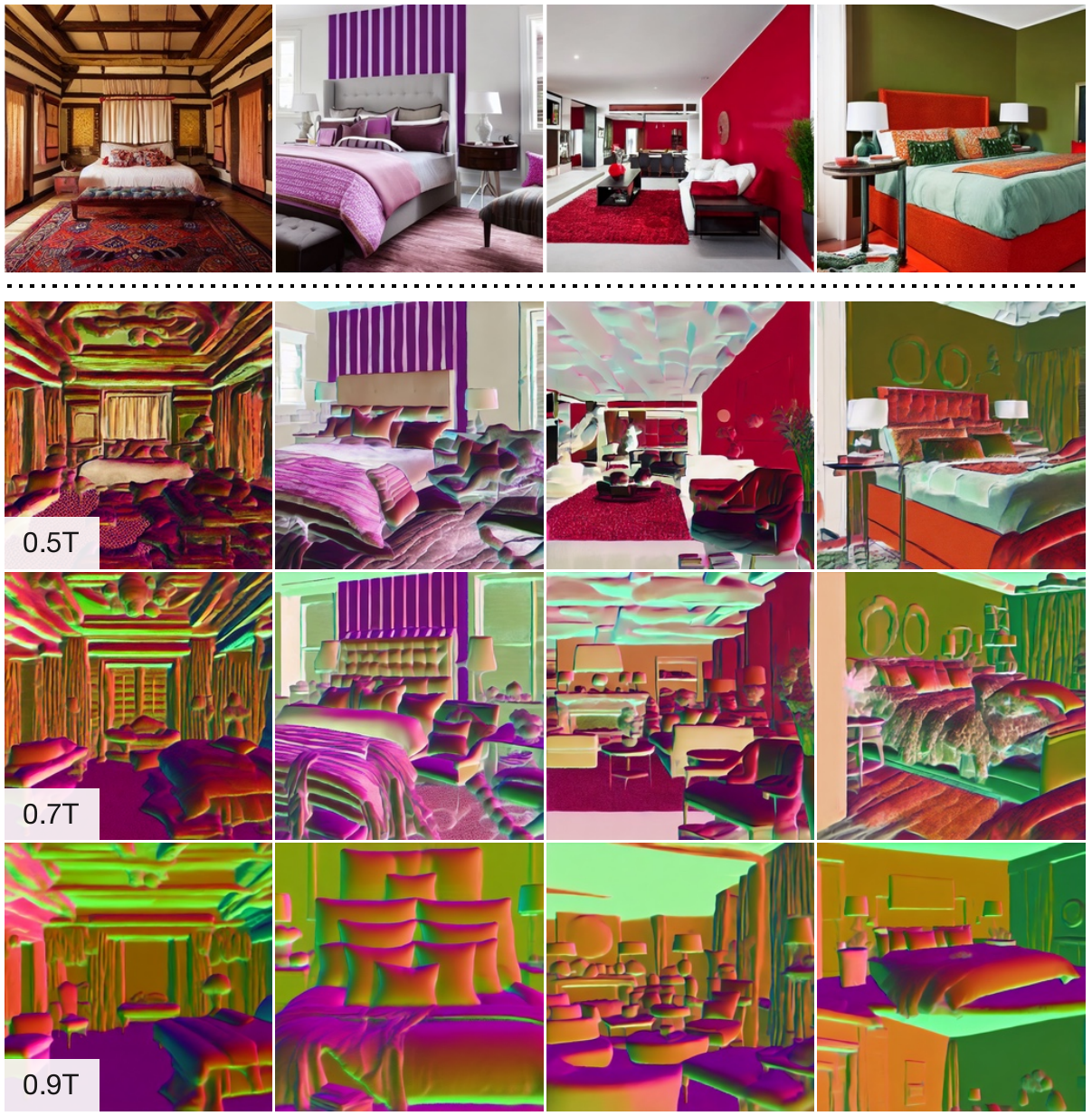}
    \caption{\textbf{Qualitative comparisons of SDEdit with starting from different time steps.} A trade-off exists between the effect of style transfer and content preservation.}
    \label{fig:sdedit-step}
\end{figure}
\begin{table}[t] \footnotesize
    \centering
    \caption{\textbf{Quantitative comparisons on in-domain surface normal estimation between different timesteps where the generation process of SDEdit starts.} The performance improves at the expense of deviation from input image contents.}
    \begin{tabular}{lcc}
    \toprule
    Step & L1$\downarrow$ & Ang$\downarrow$ \\
    \midrule
    0.5$T$ & 0.2897 & 0.5336 \\
    0.7$T$ & 0.2599 & 0.5087 \\
    0.9$T$ & 0.2059 & 0.4568 \\
    \bottomrule
    \end{tabular}
    \label{tab:sdedit-step}
\end{table}

\paragraph{SDEdit.}
The time steps from which the generation process of SDEdit starts can be seen as the strength of preserving the contents of input images. We show in \cref{fig:sdedit-step} that generating from step 0.5$T$ produces images with similar contents to the input images, while from step 0.9$T$ results in plausible estimation of surface normals, but the image contents are disrupted, despite achieving the best performance in quantitative evaluation reported in \cref{tab:sdedit-step}. This issue has long been understood as a trade-off between the effect of style transfer and content preservation in image-to-image literature~\cite{Huang_2017_ICCV, NIPS2017_49182f81}, but for deterministic dense prediction problems considered in this work, such a trade-off is not permitted.

\begin{figure}
    \centering
    \includegraphics[width=\linewidth]{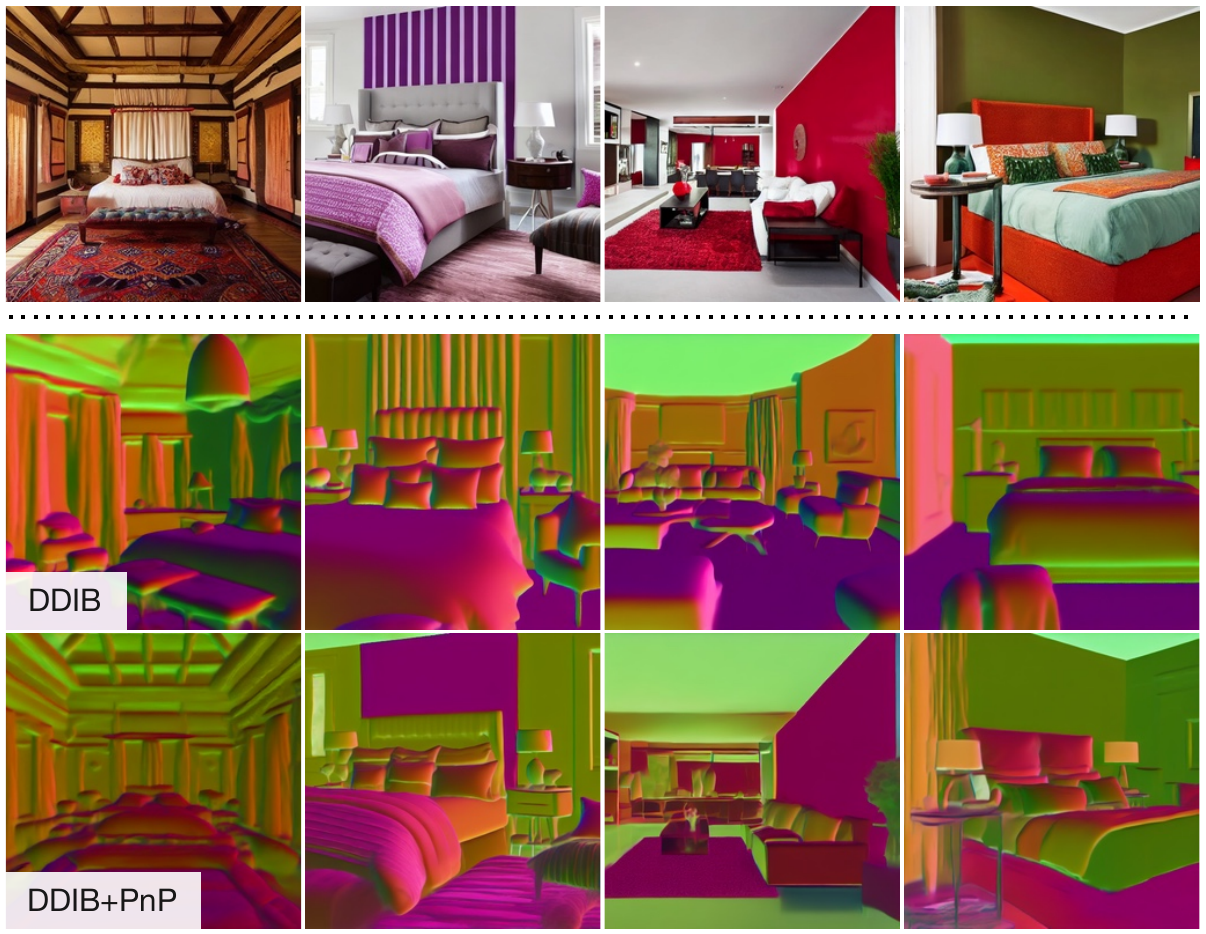}
    \caption{\textbf{Qualitative comparisons between DDIB and DDIB with Plug-and-Play (PnP).} The image contents are reserved but not consistent with accurate normals.}
    \label{fig:ddib-pnp}
\end{figure}
\begin{table}[t] 
    \centering \footnotesize
    \caption{\textbf{Quantitative comparisons on in-domain surface normal estimation between DDIB and DDIB with Plug-and-Play (PnP).} The feature injection regulates the generated contents while improving performance.}
    \begin{tabular}{lcc}
    \toprule
    Variants & L1$\downarrow$ & Ang$\downarrow$ \\
    \midrule
    DDIB & 0.1849 & 0.4210 \\
    DDIB + PnP & 0.1652 & 0.3634 \\
    \bottomrule
    \end{tabular}
    \label{tab:ddib-pnp}
\end{table}

\paragraph{DDIB.}
As presented in \cref{sec:baseline-result}, DDIB is capable of generating images that are likely sampled from output distributions, but the contents and geometry of output results are not consistent with input images. We explore the approach to content consistency by adopting feature constraints proposed by Plug-and-Play (PnP)~\cite{Tumanyan_2023_CVPR} for image-to-image translation, which injects the self-attention and convolution features of input images into output images. As shown in \cref{fig:ddib-pnp} and \cref{tab:ddib-pnp}, the structures and contents of output images of DDIB with PnP constraints highly resemble the input images, but the estimated normals remain inaccurate despite better quantitative performance.

\begin{figure}
    \centering
    \includegraphics[width=\linewidth]{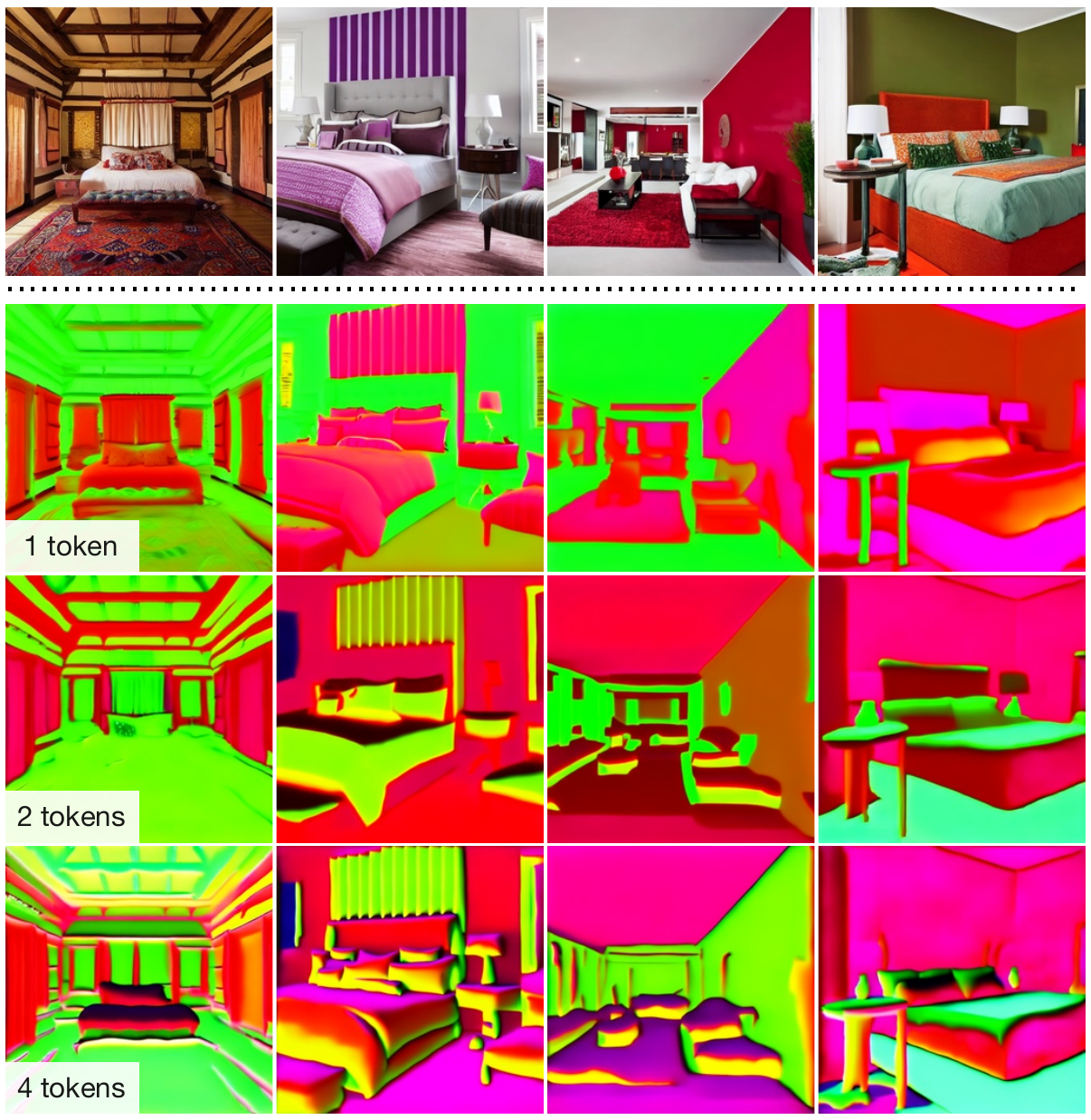}
    \caption{Qualitative comparisons of IP2P with different training tokens.}
    \label{fig:ip2p-token}
\end{figure}

\paragraph{IP2P.}
We analyze the expressiveness of inverted tokens by varying the number of training tokens in IP2P (learned). While the performance is slightly improved in one metric of quantitative evaluation, shown in \cref{tab:ip2p-token}, \cref{fig:ip2p-token} reveals that the differences between the estimated results are not significant.

\begin{table}[t] 
    \caption{\textbf{Quantitative comparisons on in-domain surface normal estimation between different training tokens of IP2P (learned).} The increased number of tokens does not guarantee improved performance.}
    \centering \footnotesize
    \begin{tabular}{lcc}
    \toprule
    \#Tokens & L1$\downarrow$ & Ang$\downarrow$ \\
    \midrule
    1 & 0.3550 & 0.7181 \\
    2 & 0.3470 & 0.7790 \\
    4 & 0.3274 & 0.6384 \\
    \bottomrule
    \end{tabular}
    \label{tab:ip2p-token}
\end{table}

\subsection{Failure Cases}

\begin{figure}
    \centering
    \includegraphics[width=\linewidth]{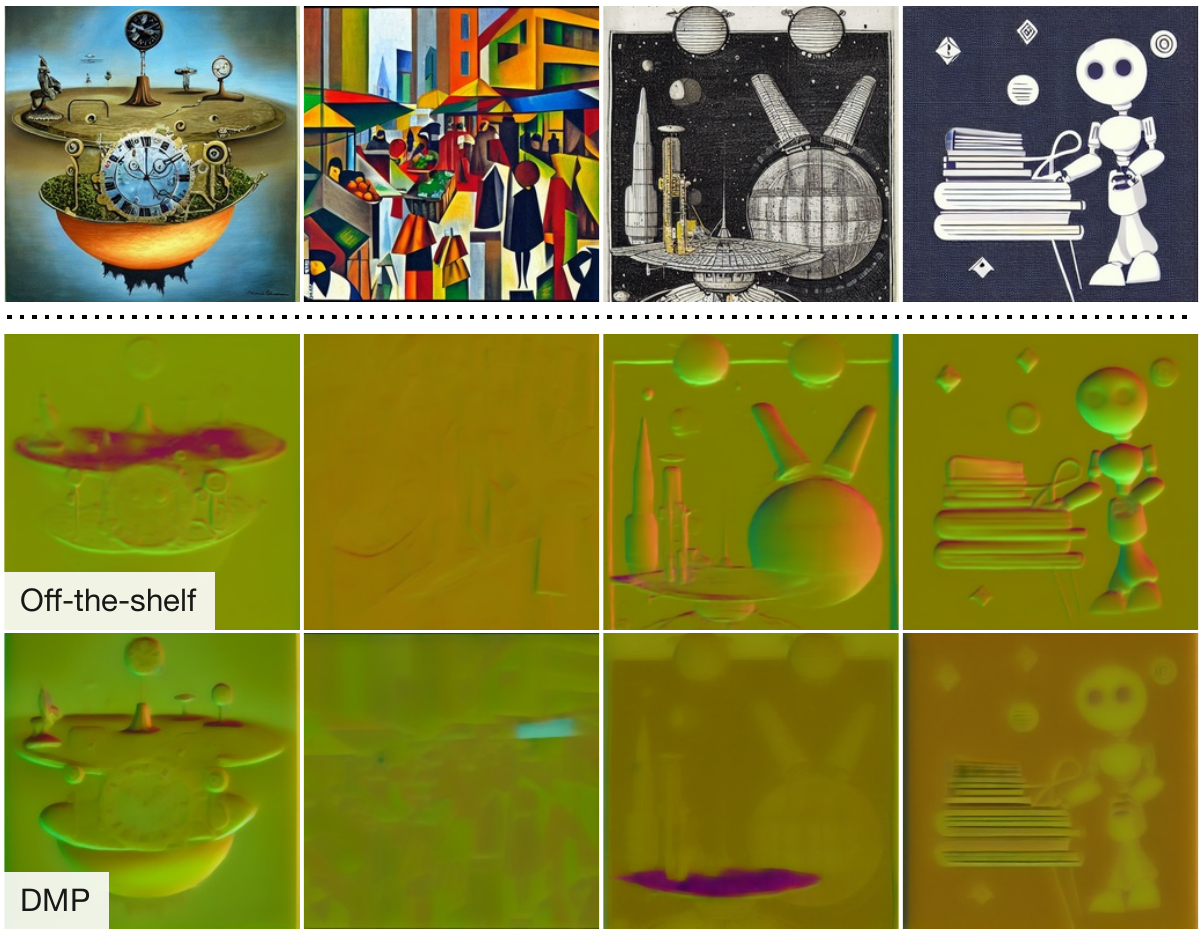}
    \caption{Failure cases of surface normal estimation.}
    \label{fig:fail-normal}
\end{figure}

\begin{figure}
    \centering
    \includegraphics[width=\linewidth]{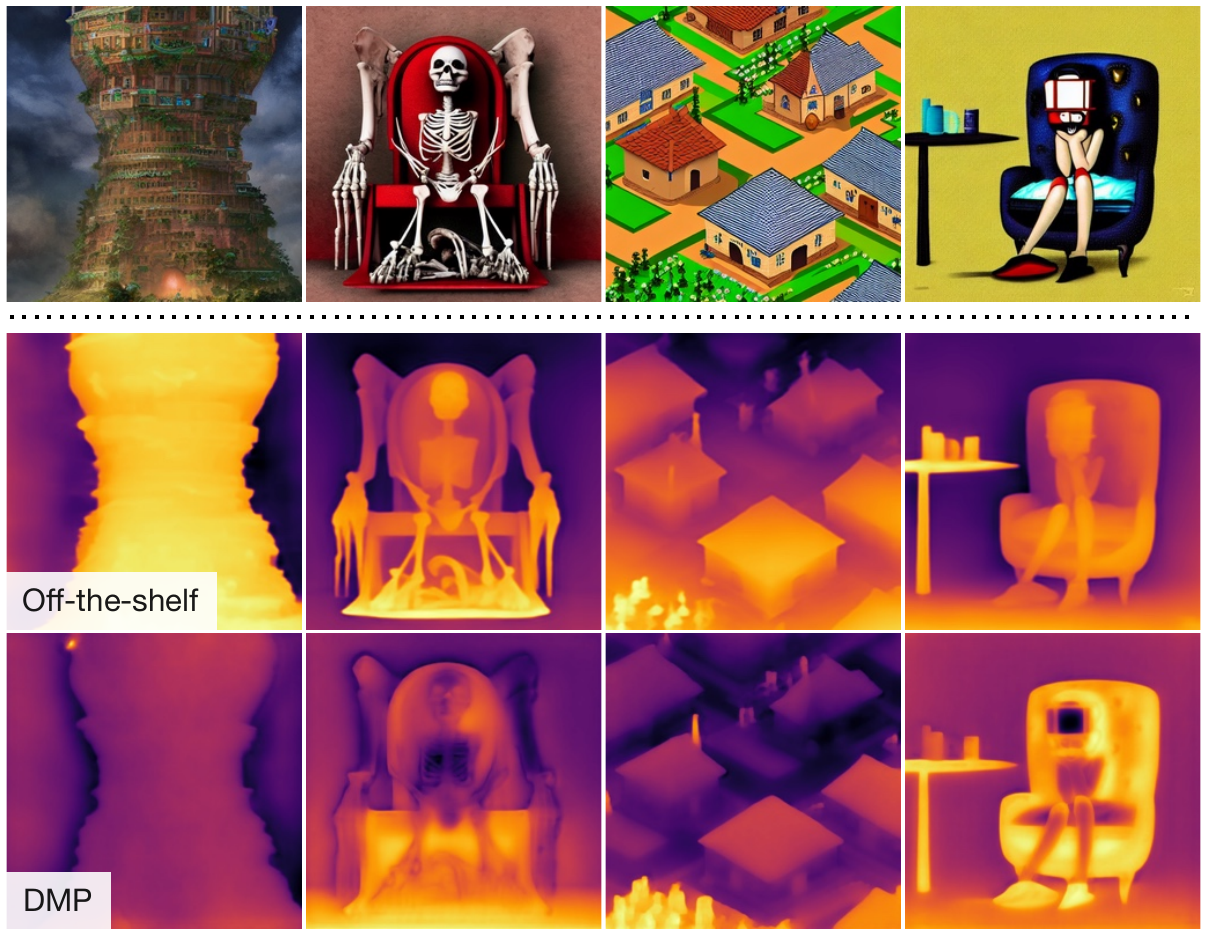}
    \caption{Failure cases of depth estimation.}
    \label{fig:fail-depth}
\end{figure}

\begin{figure}
    \centering
    \includegraphics[width=\linewidth]{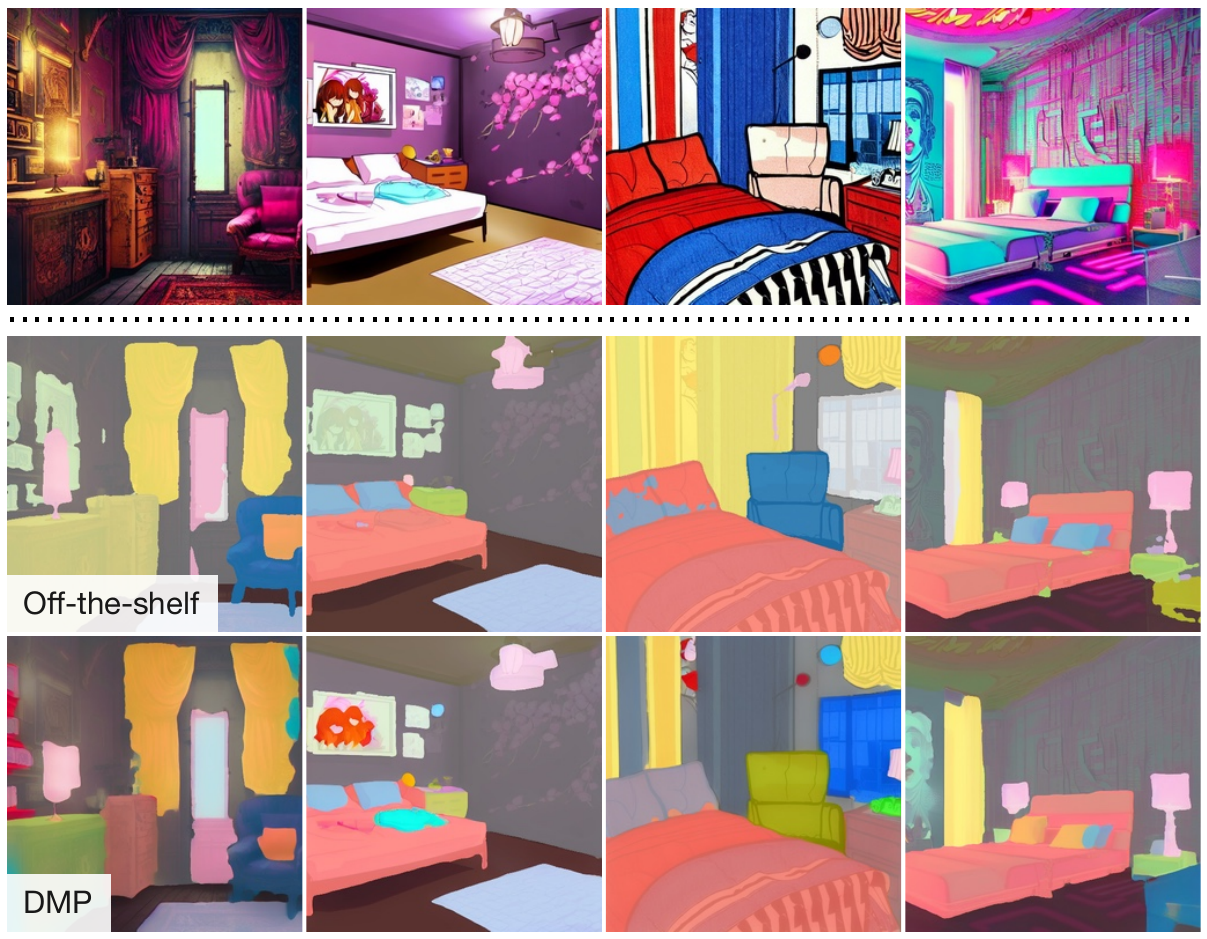}
    \caption{Failure cases of semantic segmentation.}
    \label{fig:fail-seg}
\end{figure}

We demonstrate some examples of failure cases in \cref{fig:fail-normal} for surface normal estimation, \cref{fig:fail-depth} for depth estimation, and \cref{fig:fail-seg} for semantic segmentation, where off-the-shelf approaches might provide more accurate prediction than our method.

\begin{table*}[t]
    \caption{\textbf{Style templates}, where \{\} is replaced by original prompts.}
    \centering
    \hrule
    \vspace{2mm}
    \begin{compactitem}
        \item anime artwork, \{\} . anime style, key visual, vibrant, studio anime, highly detailed
        \item concept art, \{\} . digital artwork, illustrative, painterly, matte painting, highly detailed
        \item comic, \{\} . graphic illustration, comic art, graphic novel art, vibrant, highly detailed
        \item neonpunk style, \{\} . cyberpunk, vaporwave, neon, vibes, vibrant, stunningly beautiful, crisp, detailed, sleek, ultramodern, magenta highlights, dark purple shadows, high contrast, cinematic, ultra-detailed, intricate, professional \\
        surrealist art, \{\} . dreamlike, mysterious, provocative, symbolic, intricate, detailed
        \item abstract style, \{\} . non-representational, colors and shapes, expression of feelings, imaginative, highly detailed
        \item art deco style, \{\} . geometric shapes, bold colors, luxurious, elegant, decorative, symmetrical, ornate, detailed
        \item vaporwave style, \{\} . retro aesthetic, cyberpunk, vibrant, neon colors, vintage 80s and 90s style, highly detailed 
    \end{compactitem}
    \vspace{2mm}
    \hrule
    \label{tab:styles}
\end{table*}

\subsection{Results of Compared Methods}\label{sec:baseline-result}

\begin{figure*}
    \centering
    \includegraphics[width=0.83\linewidth]{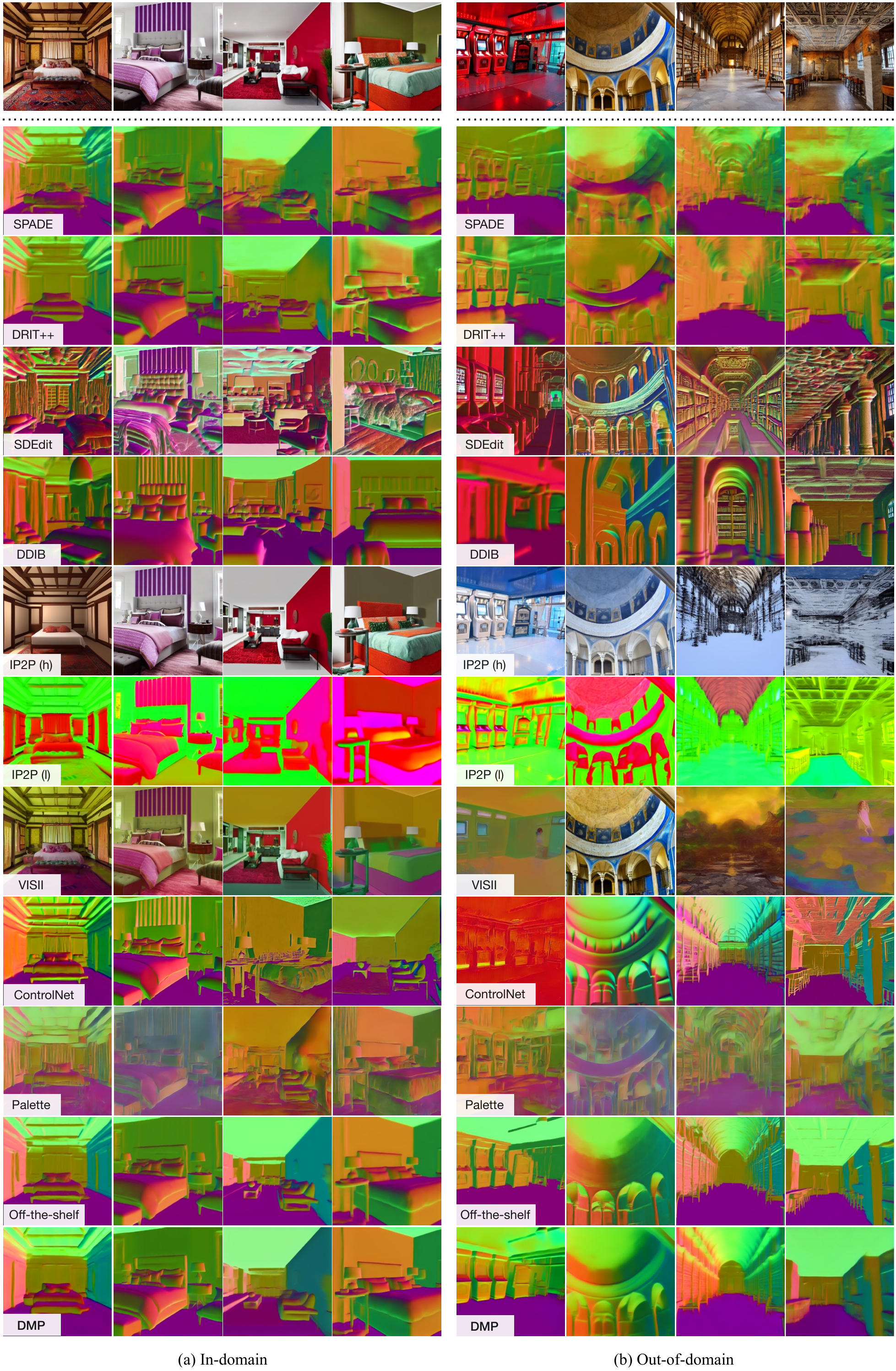}
    \vspace{-2mm}
    \caption{Qualitative results of compared methods on surface normal estimation.}
    \label{fig:baselines-normal}
\end{figure*}

\begin{figure*}
    \centering
    \includegraphics[width=0.83\linewidth]{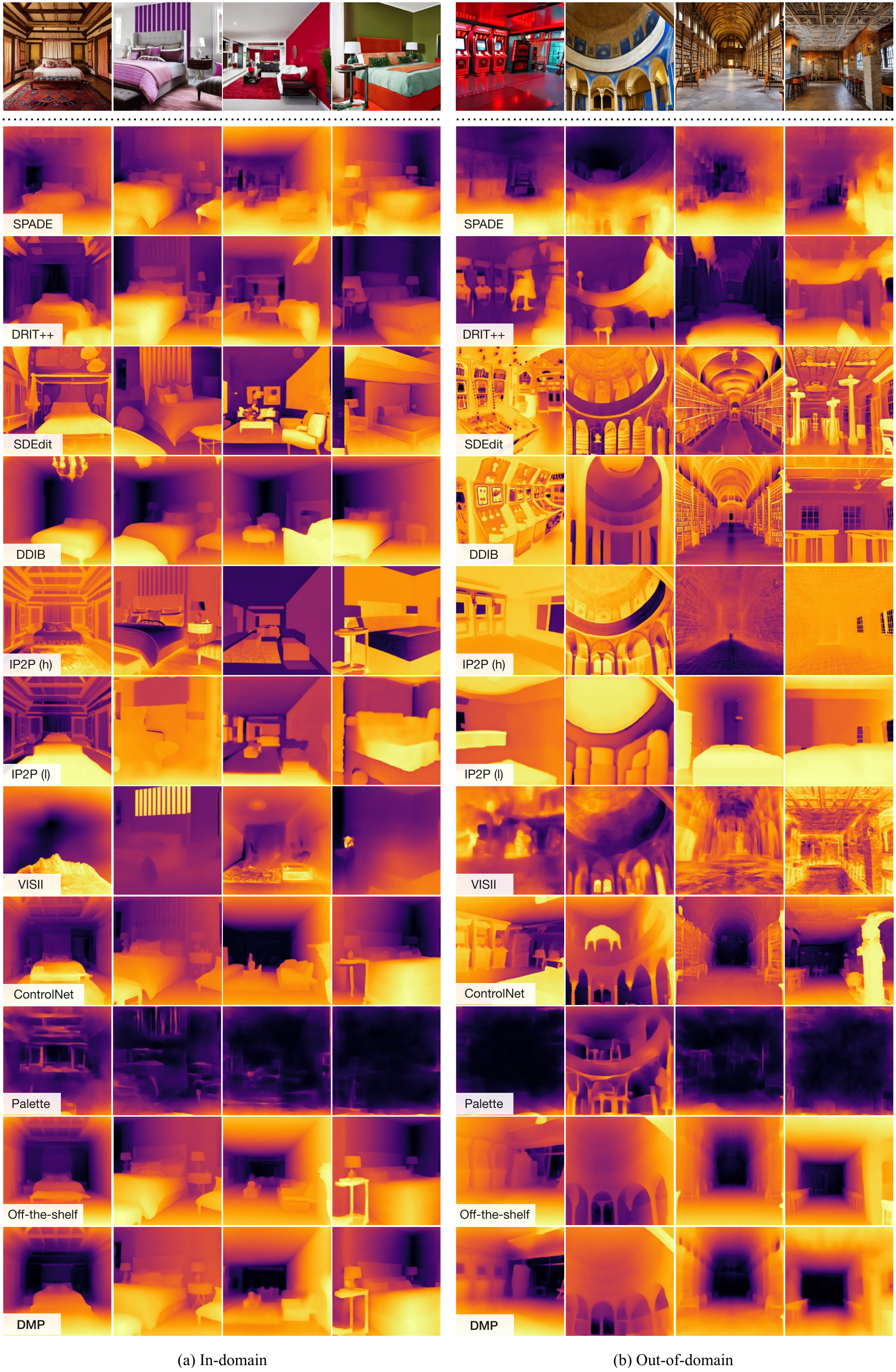}
    \vspace{-2mm}
    \caption{Qualitative results of compared methods on depth estimation.}
    \label{fig:baselines-depth}
\end{figure*}

\begin{figure*}
    \centering
    \includegraphics[width=0.83\linewidth]{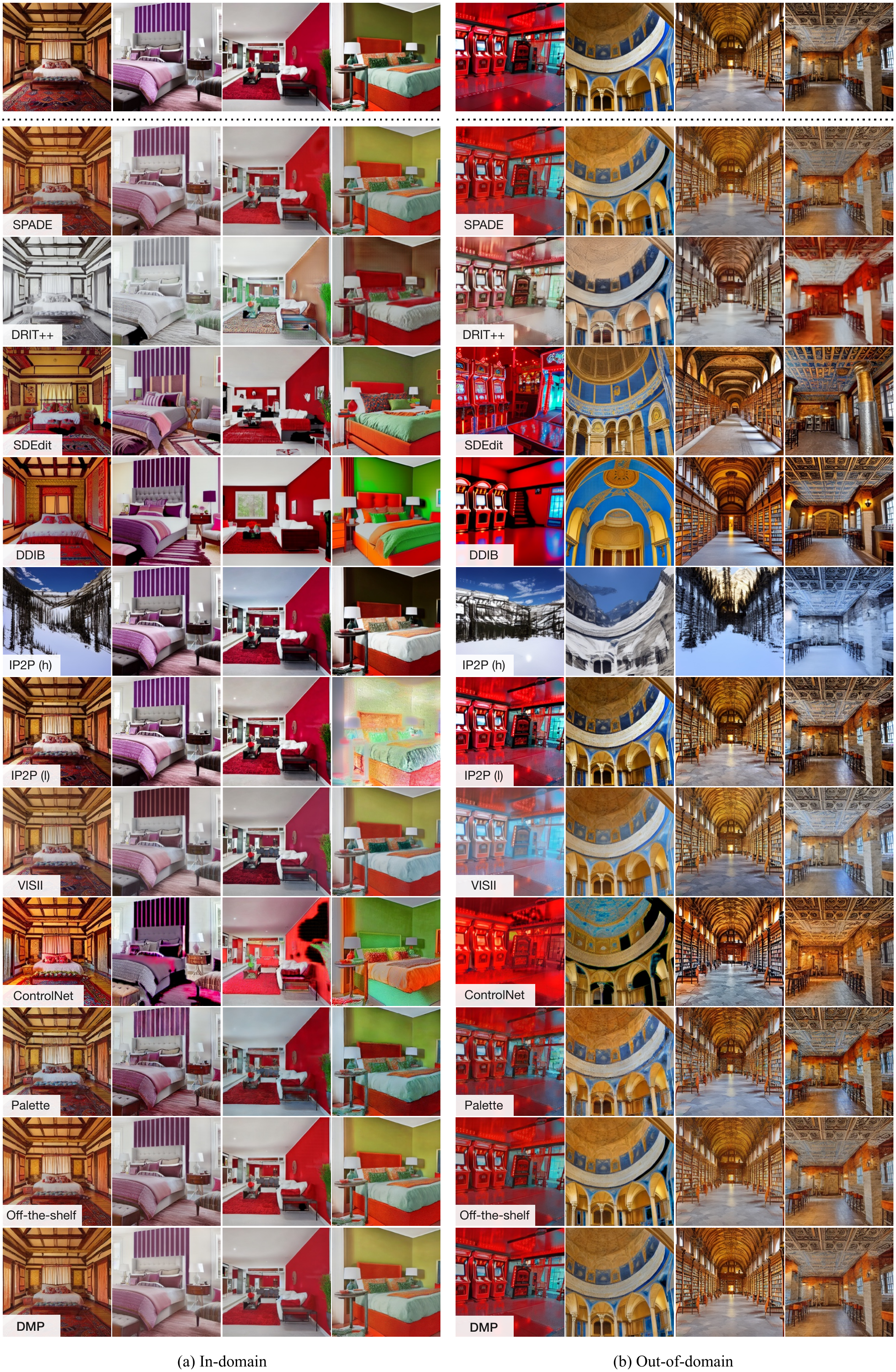}
    \vspace{-2mm}
    \caption{Qualitative results of compared methods on albedo estimation.}
    \label{fig:baselines-albedo}
\end{figure*}

\begin{figure*}
    \centering
    \includegraphics[width=0.83\linewidth]{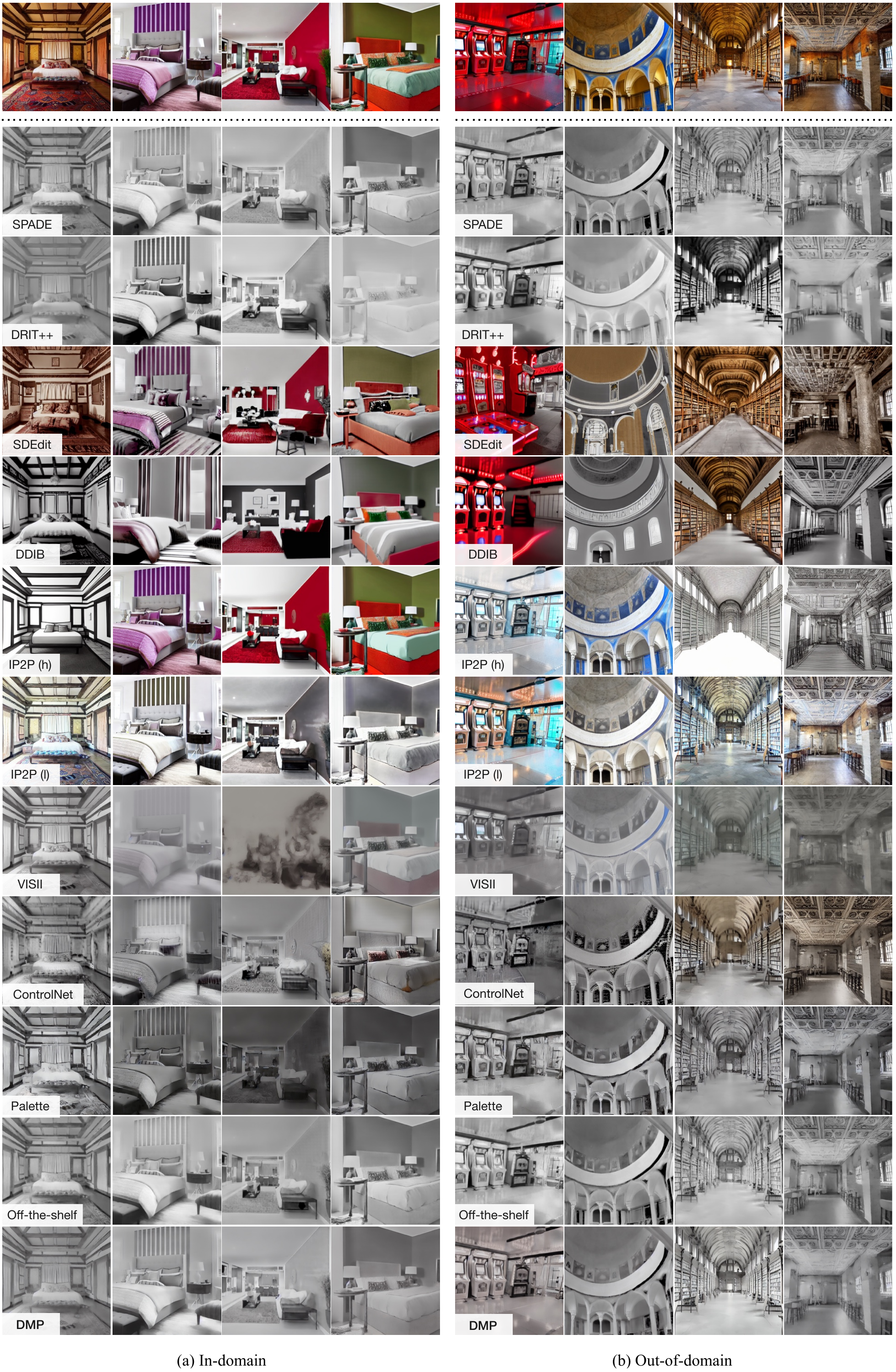}
    \vspace{-2mm}
    \caption{Qualitative results of compared methods on shading estimation.}
    \label{fig:baselines-shade}
\end{figure*}

\begin{figure}
    \centering
    \includegraphics[width=0.95\linewidth]{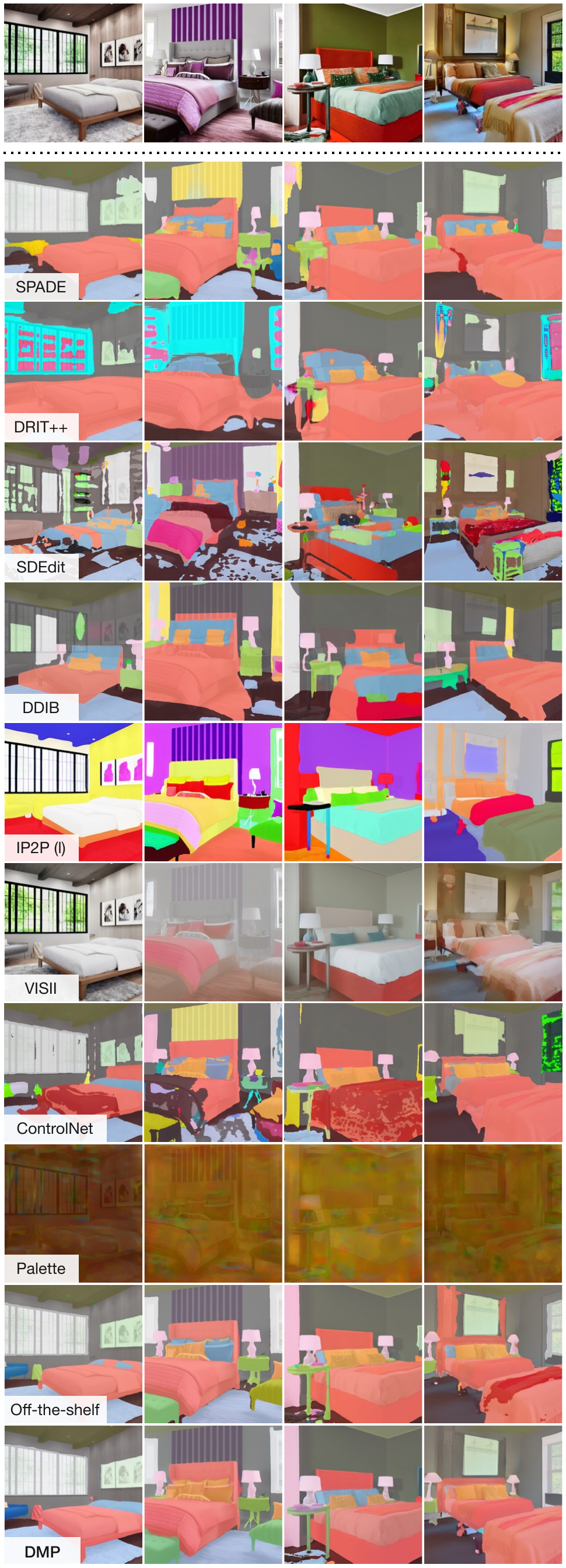}
    \vspace{-2mm}
    \caption{Qualitative results of compared methods on semantic segmentation.}
    \label{fig:baselines-seg}
\end{figure}

We show the example images generated by the compared methods listed in Section 4.1. The results of surface normal estimation are in \cref{fig:baselines-normal}, with depths in \cref{fig:baselines-depth}, albedo in \cref{fig:baselines-albedo}, shading in \cref{fig:baselines-shade}, and semantic segmentation in \cref{fig:baselines-seg}.

\vspace{-2mm}

\section{Implementation Details}

\paragraph{Model Architecture and Optimization.}
We use Stable Diffusion 1.4 as the pre-trained text-to-image model and adapt it with rank $= 4$ for LoRA. We fine-tune the model for $50$K steps with batch size $8$ and learning rate $0.0001$ with a cosine decay schedule. The training takes around 14 hours with a single NVIDIA RTX 3090.

\paragraph{Generating Images.}
We generate the training and test images by first generating a set of prompts with a large language model. The prompt for the language model is a template adapted from pix2pix-zero~\cite{pix2pixzero}, where different scene keywords are filled in. The template is
\begin{quote}
    ``Provide a caption for a photo of a \texttt{scene}. The caption should contain many adjectives, should describe colors, styles, lighting and materials in the photo, should be in English and should be no longer than 150 characters. Caption:''.
\end{quote}
The placeholder \texttt{scene} is replaced by ``bedroom'' for training images and in-domain test images. To generate out-of-domain test images for estimating 3D properties and intrinsic images, it is replaced by uniform sampling from the keywords in \cref{tab:scenes}.

Out-of-domain test images for segmentation are synthesized by varying the image styles of in-domain test images, for semantic categories should remain the same across training and test images. The prompts regulating the styles are listed in \cref{tab:styles} borrowed from an online post~\cite{106styles}.

\section{Applications}

Surface normals and depths facilitate many vision tasks. We show by the examples of 3D photo inpainting~\cite{shih20203d} that precise depths improve 3D reconstruction from 2D images. Compared to the default depth estimator~\cite{midas}, the resulting videos produced with the depth maps generated by our approach have more accurate depth relationships between the objects. Please refer to the project website for visual demonstrations.

\clearpage
\onecolumn
\begin{longtable}{llll}
    \caption{\textbf{Scenes categories} of out-of-domain images.} \\
    \midrule
    airlock & airplane cabin & airport terminal & airport ticket counter \\
    alcove & amusement arcade & anechoic chamber & indoor apse \\
    aquarium & arcade & archive & armory \\
    indoor arrival gate & art gallery & art school & art studio \\
    artists loft & assembly line & indoor athletic field & attic \\
    auditorium & auto factory & indoor auto mechanics & auto showroom \\
    backstage & indoor badminton court & baggage claim & ball pit \\
    ballroom & indoor bank & bank vault & banquet hall \\
    indoor baptistry & bar & barbershop & barrack \\
    basement & indoor basketball court & bathhouse & bathroom \\
    indoor batting cage & indoor bazaar & beauty salon & bedchamber \\
    bedroom & beer hall & belfry & bell foundry \\
    berth & berth deck & betting shop & bicycle racks \\
    bindery & biology laboratory & indoor bistro & indoor bleachers \\
    indoor bomb shelter & bookbindery & bookstore & indoor booth \\
    indoor bow window & bowling alley & box seat & boxing ring \\
    breakroom & indoor brewery & indoor brickyard & burial chamber \\
    indoor bus depot & bus interior & indoor bus station & butchers shop \\
    indoor cabin & cafeteria & call center & candy store \\
    canteen & backseat car interior & frontseat car interior & cardroom \\
    cargo container interior & indoor carport & indoor casino & catacomb \\
    indoor cathedral & catwalk & chapel & checkout counter \\
    cheese factory & chemistry lab & indoor chicken coop & indoor chicken farm \\
    childs room & interior choir loft & indoor church & indoor circus tent \\
    classroom & clean room & indoor clock tower & indoor cloister \\
    closet & clothing store & cockpit & coffee shop \\
    computer room & conference center & conference hall & conference room \\
    confessional & control room & indoor control tower & indoor convenience store \\
    corridor & courtroom & interior covered bridge & crawl space \\
    cybercafe & indoor dairy & dance school & darkroom \\
    day care center & delicatessen & dentists office & department store \\
    departure lounge & indoor diner & dining car & dining hall \\
    dining room & discotheque & distillery & indoor doorway \\
    dorm room & dress shop & dressing room & indoor driving range \\
    drugstore & editing room & electrical room & elevated catwalk \\
    interior elevator & elevator lobby & elevator shaft & engine room \\
    entrance hall & indoor escalator & exhibition hall & fabric store \\
    indoor factory & fastfood restaurant & indoor ferryboat & indoor firing range \\
    fishmarket & interior fitting room & indoor flea market & indoor florist shop \\
    food court & indoor foundry & funeral chapel & funeral home \\
    furnace room & galley & game room & indoor garage \\
    indoor general store & indoor geodesic dome & gift shop & great hall \\
    indoor greenhouse & indoor gun deck & gun store & indoor gymnasium \\
    hallway & indoor hangar & hardware store & hat shop \\
    hatchery & hatchway & hayloft & hearth \\
    home office & home theater & hospital room & indoor hot tub \\
    hotel breakfast area & hotel room & indoor hunting lodge & ice cream parlor \\
    indoor ice skating rink & indoor incinerator & indoor inn & indoor jacuzzi \\
    indoor jail & jail cell & jewelry shop & jury box \\
    indoor kennel & kindergarden classroom & indoor kiosk & kitchen \\
    kitchenette & lab classroom & indoor labyrinth & landing \\
    laundromat & lavatory & lecture room & legislative chamber \\
    indoor library & indoor lido deck & limousine interior & indoor liquor store \\
    living room & lobby & locker room & loft \\
    indoor lookout station & indoor lumberyard & machine shop & indoor market \\
    martial arts gym & maternity ward & mess hall & mezzanine \\
    military hospital & mill & mine & indoor mini golf course \\
    indoor monastery & morgue & indoor mosque & indoor movie theater \\
    indoor museum & music store & music studio & natural history museum \\
    newsroom & indoor newsstand & nightclub & indoor nuclear power plant \\
    nursery & nursing home & indoor observatory & office \\
    office cubicles & indoor oil refinery & operating room & optician \\
    orchestra pit & interior organ loft & orlop deck & ossuary \\
    indoor outhouse & oyster bar & packaging plant & palace hall \\
    pantry & paper mill & indoor parking garage & parlor \\
    particle accelerator & indoor party tent & pawnshop & penalty box \\
    perfume shop & pet shop & pharmacy & physics laboratory \\
    piano store & pig farm & indoor pilothouse & pizzeria \\
    indoor planetarium & playroom & indoor podium & portrait studio \\
    indoor power plant & print shop & promenade deck & indoor pub \\
    pulpit & pump room & indoor quonset hut & reading room \\
    reception & recreation room & indoor recycling plant & refectory \\
    repair shop & restaurant & restaurant kitchen & indoor restroom \\
    revolving door & riding arena & indoor roller skating rink & rolling mill \\
    sacristy & sauna & sawmill & science museum \\
    scriptorium & security check point & server room & sewer \\
    sewing room & shipping room & indoor shipyard & shoe shop \\
    indoor shopping mall & shower & shower room & shrine \\
    indoor skywalk & sporting goods store & squash court & stable \\
    indoor stage & staircase & indoor steam plant & indoor steel mill \\
    storage room & storeroom food & submarine interior & subway interior \\
    supermarket & sushi bar & indoor swimming pool & indoor synagogue \\
    tearoom & teashop & television room & television studio \\
    indoor tennis court & indoor tent & textile mill & indoor procenium theater \\
    indoor round theater & indoor seats theater & thriftshop & throne room \\
    ticket booth & indoor ticket window & indoor tobacco shop & toyshop \\
    indoor track & trading floor & train interior & rail indoor tunnel \\
    road indoor tunnel & turkish bath & utility room & utility tunnel \\
    van interior & indoor velodrome & ventilation shaft & vestry \\
    veterinarians office & videostore & indoor volleyball court & voting booth \\
    waiting room & walk in freezer & indoor warehouse & indoor washhouse \\
    indoor water treatment plant & wet bar & whispering gallery & wig shop \\
    window seat & winery & witness stand & workroom \\
    workshop & indoor wrestling ring & youth hostel & basketball arena \\
    football arena & hockey arena & performance arena & rodeo arena \\
    soccer arena & home atrium & public atrium & bakery kitchen \\
    bakery shop & airplane cargo deck & choir loft & cloakroom booth \\
    cloakroom & library cubicle & office cubicle & home dinette \\
    vehicle dinette & elevator door & freight elevator & ferryboat cargo deck \\
    fitting room & organ loft & establishment poolroom & home poolroom \\
    spa massage room & spa mineral bath & corridor in a subway station & platform in a subway station \\
    turnstiles in a subway station & platform in a train station & station in a train station & barrel storage in a wine cellar \\
    bottle storage in a wine cellar \\
    \midrule
    \label{tab:scenes}
\end{longtable}
\twocolumn

\end{document}